\documentclass[11pt]{article}

\usepackage[preprint]{acl}

\usepackage{times}
\usepackage{latexsym}
\usepackage{amsmath}

\usepackage{pgfplots}
\usepackage{tikz}
\pgfplotsset{compat=1.18}
\definecolor{myblue}{RGB}{76,114,176}
\definecolor{myorange}{RGB}{221,132,82}
\definecolor{mygreen}{RGB}{34,139,34}   
\definecolor{myred}{RGB}{178,34,34}     
\definecolor{mypurple}{RGB}{129,114,177}
\definecolor{myteal}{RGB}{0,128,160}
\definecolor{headergray}{RGB}{230,236,243}   
\definecolor{groupgray}{RGB}{214,223,233}    
\definecolor{lowgreen}{RGB}{223,239,216}     
\definecolor{midyellow}{RGB}{250,239,210}    
\definecolor{highred}{RGB}{244,214,214}      
\usepackage[table]{xcolor}
\newcommand{\std}[1]{\hspace{0.08em}{\scriptsize$\pm$#1}}
\newcommand{\overall}[2]{\cellcolor{lowgreen}\textbf{#1\std{#2}}}

\usepackage[T1]{fontenc}

\usepackage[utf8]{inputenc}

\usepackage{microtype}
\usepackage{float}

\usepackage{inconsolata}

\usepackage{lipsum}

\usepackage{graphicx}

\usepackage[normalem]{ulem} 
\usepackage{booktabs}
\usepackage{multirow}
\usepackage{amssymb}
\usepackage{ragged2e}

\usepackage{makecell}
\usepackage{placeins}

\usepackage{comment}
\usepackage{changepage}

\usepackage{hyperref} 

\usepackage[most]{tcolorbox}
\usepackage{array}
\usepackage{colortbl}

\definecolor{promptbg}{RGB}{245,246,247}
\definecolor{promptborder}{RGB}{200,200,200}

\usepackage{titlesec}
\usepackage{subcaption}

\usepackage{pifont}
\usepackage{enumitem}
\definecolor{lightpurple}{RGB}{238,231,246}
\definecolor{lightyellow}{RGB}{255, 247, 204}
\usepackage{enumitem}

\usepackage{ulem}
\usepackage{tabularx}

\definecolor{bestgreen}{RGB}{226,239,218}

\newcommand{\best}[1]{\cellcolor{green!10}\textbf{#1}}

\titlespacing*{\section}{0pt}{1.2ex plus 0.2ex minus 0.2ex}{0.6ex}
\titlespacing*{\subsection}{0pt}{0.9ex plus 0.2ex minus 0.2ex}{0.4ex}
\titlespacing*{\subsubsection}{0pt}{0.7ex plus 0.1ex minus 0.1ex}{0.3ex}

\newif\ifdraft
\draftfalse

\usepackage{xcolor}

\ifdraft
  \usepackage{CJKutf8}
\fi

\ifdraft

\newcommand{\bobo}[1]{{\color{red}LiBobo: #1}}

\newcommand{\jiajia}[1]{{\color{blue}SongJiaJia: #1}}
\else

  \newcommand{\bobo}[1]{}
  \newcommand{\jiajia}[1]{}
\fi

\newcommand{\startCJK}{\ifdraft\begin{CJK*}{UTF8}{gbsn}\fi}
\newcommand{\stopCJK}{\ifdraft\end{CJK*}\fi}

\title{
From Profiling to Synthesis: Benchmarking Implicit Behavioral Alignment in Personalized LLM Agents
}

\author{
\textbf{Jiajia Song}$^{1}$ \quad
\textbf{Bobo Li}$^{1}$ \quad
\textbf{Haiwen Yi}$^{2}$ \quad
\textbf{Zibo Ji}$^{3}$ \quad
\textbf{Meishan Zhang}$^{4}$ \\
\textbf{Hao Fei}$^{5}$ \quad
\textbf{Min Zhang}$^{4}$ \quad
\textbf{Mong-Li Lee}$^{1}$ \quad
\textbf{Wynne Hsu}$^{1}$ \\[3pt]
$^{1}$National University of Singapore \quad
$^{2}$University of Toronto \\
$^{3}$University of Minnesota Twin Cities \quad
$^{4}$Harbin Institute of Technology, Shenzhen \\
$^{5}$University of Oxford \\[3pt]
\texttt{\{jjiasong,libobo,dcsleeml,dcshsuw\}@nus.edu.sg},
\texttt{  hao.fei@bdi.ox.ac.uk}
}

\begin{document}
\startCJK
\maketitle

\begin{abstract}
Large Language Models have enabled increasingly capable autonomous agents, yet personalization remains critical for making such  agents  practically useful.
Recent benchmarks have begun evaluating personalization in agents, but they largely rely on static preference snapshots, fixed interaction logs, or question answering over predefined user profiles. Such designs fail to capture the complexity of evolving user preferences and neglect preference-conditioned task execution—a discrepancy we term as the \emph{knowledge-to-action  gap}.
To address this challenge, we introduce \textsc{IBA-Bench}, a benchmark for implicit behavioral alignment constructed from longitudinal interaction histories that contain noise, implicit cues, and temporal inconsistencies. Unlike prior work,
\textsc{IBA-Bench} evaluates whether an agent can execute tasks while satisfying implicit user constraints inferred from historical interactions.
We further propose \textsc{IBA-Agent}, an agent framework that reconciles conflicting priorities  through broad retrieval and trajectory-level alignment.
Experiment results on \textsc{IBA-Bench} show that effective personalization remains a significant  challenge for state-of-the-art LLM agents, and the proposed  \textsc{IBA-Agent} substantially improves behavioral alignment in complex scenarios across nine application domains.
\end{abstract}

\section{Introduction}

\begin{figure*}[t]
    \centering
    \includegraphics[width=0.85\textwidth]{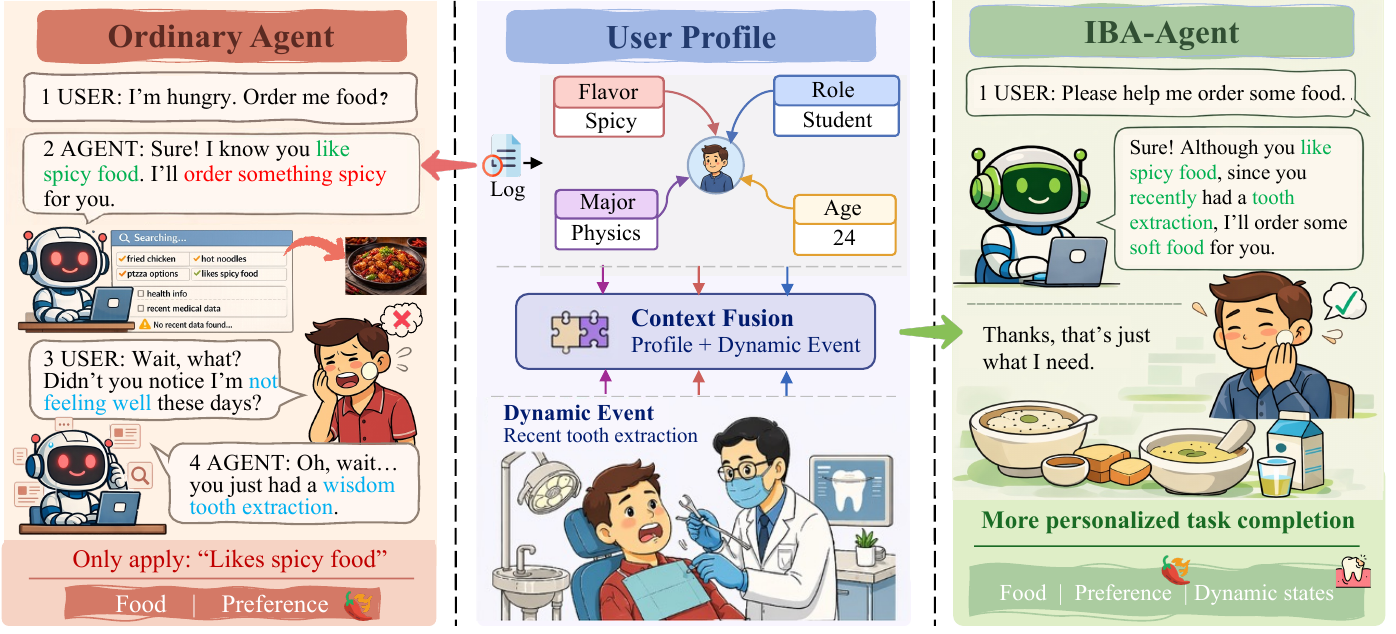}
    \caption{Personalization beyond QA:  knowledge-to-action gap between preference inference and task execution.
    }
    \vspace{-5mm}
    \label{fig:fig4}
\end{figure*}

Large Language Models (LLMs)~\cite{openai2024gpt4technicalreport, comanici2025gemini, guo2025DeepSeek} have evolved from conversational chatbots into autonomous agents capable of executing complex real-world tasks~\cite{yang18hotpotqa, yang24sweagent}.
This shift makes personalization essential as users now expect  assistants that can adapt to their  preferences and constraints
rather than simply generic tool executors~\cite{samuel24personagym}. Achieving such personalization is a challenge because user preferences are 
rarely explicit; instead, they dynamically evolve and remain implicitly embedded within lengthy interaction histories~\cite{zhang25personaagent}. Capturing and synthesizing these dynamic signals across long interaction histories
 is needed  for building the next-generation personalized agents.

Recent works have introduced various personalized agents and evaluation frameworks. Early efforts extracted static profiles for response generation~\cite{joko24dpl}, collapsing preferences into flat tags and ignoring temporal dynamics. Subsequent benchmarks incorporated raw dialogue histories to preserve realism. For instance, HiCUPID~\cite{mok25etp} retains conversational trajectories, while PersonaBench~\cite{tan25personabench} models dynamic attribute evolution. Task-oriented evaluations like PersonaLens~\cite{zhao25personalens} 
replace interaction histories with summarized abstracts.
However, all these evaluations predominantly rely on extraction-based Question Answering (QA). 
As a result, they fail to adequately measure an agent’s ability to integrate implicit, evolving user preferences into complex actions and decisions. We refer to this limitation as the \emph{knowledge-to-action gap}.
This gap highlights a stark disconnect: an agent may possess crucial knowledge yet fail to apply it during task execution.

 Consider a user with a historical preference for spicy food. Recent interaction history  reveals that the user has undergone a wisdom tooth extraction (Figure~\ref{fig:fig4}). When asked to order dinner, current agents typically order a spicy meal based on the  static user preference profile. Yet, the same agent is able to recall that the user's   tooth extraction indicating that it possesses the knowledge. 
 Since the dental event and  dining request seems unrelated, 
 agents frequently overlook this connection  unless explicitly prompted. An ideal personalized agent would implicitly synthesize these signals and proactively recommend soft food. Bridging this gap requires moving beyond static profiling to dynamic synthesis, where agents infer implicit constraints and reconcile conflicts autonomously.

In this work, we propose \textbf{\textsc{IBA-Bench}}, a benchmark specifically designed to assess implicit behavioral alignment in  agents. The primary objective of \textsc{IBA-Bench} is to evaluate and bridge the knowledge-to-action gap. At the data level, we construct longitudinal interaction histories containing  implicit preference cues, dynamic attributes, and  conflicts between long-term traits and short-term states. At the task level, we move beyond explicit QA and simple preference matching by providing agents with  full interaction records and the user request. The agent  carries out the request and we  evaluate  whether the request is successfully executed and that 
 the  implicit constraints inferred from the interaction history are satisfied.

We also introduce \textbf{IBA-Agent}, a personalized agent framework to bridge the gap between historical preference understanding and  active task execution. IBA-Agent consists of two key components: (i) Deep Retrieval, which isolates relevant behavioral cues from noisy interaction histories, and (ii) Broad Thinking and Deep Alignment, which synthesizes these signals
 into actionable constraints and  planning logic prior to generation. 
 
 Extensive experiments on \textsc{IBA-Bench} reveal that standalone LLMs~\cite{openai2024gpt4technicalreport, guo2025DeepSeek} and even strong general-purpose agents such as Claude Code and Hermes Agent struggle significantly on these synthesis-heavy execution tasks. 
These results empirically validate the \emph{knowledge-to-action gap}. 
 In contrast, IBA-Agent substantially improves behavioral alignment in complex scenarios, demonstrating the effectiveness of explicit trajectory-level synthesis. Our main contributions are:

\begin{itemize}[topsep=2pt,itemsep=2pt,parsep=0pt]
\item We advocate  a shift in personalized agent research from static preference profiling to dynamic preference synthesis, and  identify the knowledge-to-action
gap as a key bottleneck in translating inferred knowledge into effective task execution.%
\item We introduce \textsc{IBA-Bench}, a benchmark that evaluates implicit behavioral alignment through concrete task execution in realistic settings characterized by 
 dynamic, conflicting, and noisy contexts.
\item We propose IBA-Agent, a personalized agent framework that combines deep retrieval with trajectory-level synthesis, providing a strong empirical baseline for building agents capable of personalized reasoning.
\end{itemize}

\section{Related Work}

\noindent\textbf{LLM-Based Agents}
have emerged as an active research area
with the rapid advancement of LLMs' reasoning~\cite{wei2022chain}, planning~\cite{valmeekam2023planning}, and tool use~\cite{schick2023toolformer, li2026for}. 
Compared to generic agents, these agents seek to operate in dynamic and diverse interaction scenarios by aligning decision-making with user-specific characteristics~\cite{wang2024aipersonalifelongpersonalization}.

\begin{figure*}[t!]
    \centering
    \includegraphics[width=\textwidth]{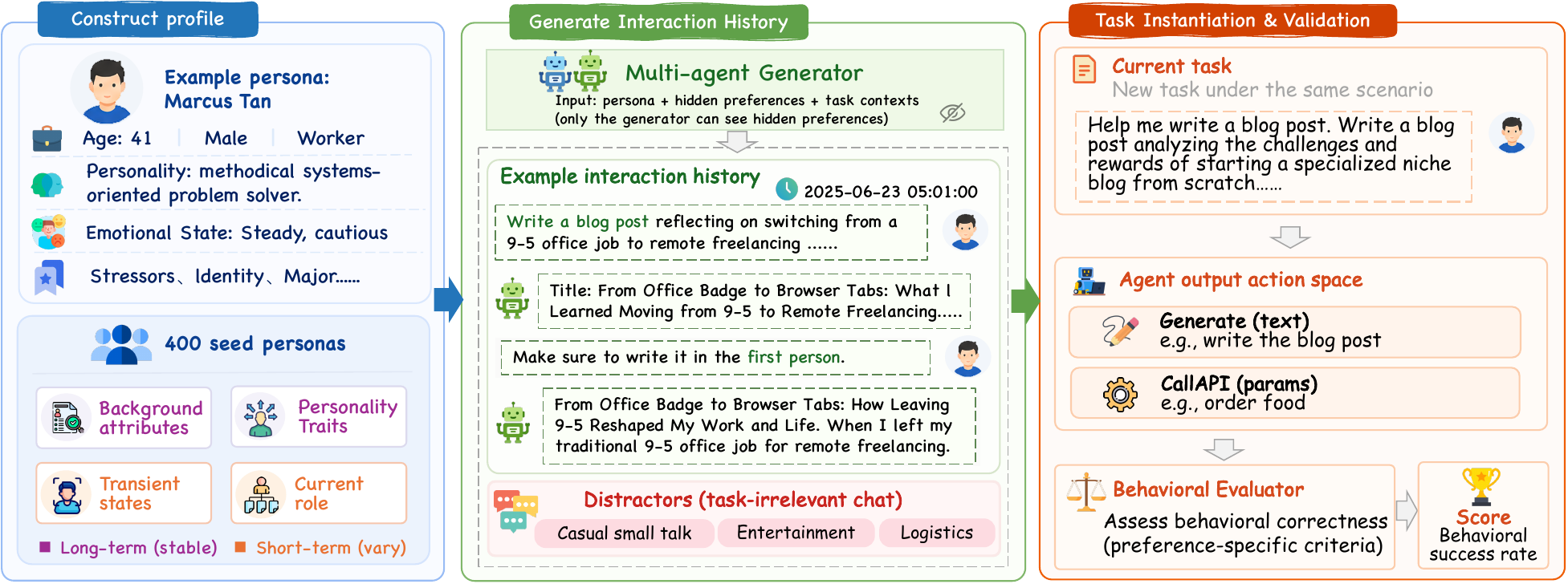}
    \caption{Benchmark construction flow chart.
    }
  
    \label{fig:fig1}
\end{figure*}

Current approaches typically introduce personalization through explicit user modeling~\cite{zhang25personaagent, afz24corr}, memory or retrieval-augmented mechanisms~\cite{ram2023context}, and adaptation based on user feedback~\cite{stiennon2022learningsummarizehumanfeedback}
 with the goal of capturing both long-term user preferences and evolving interaction contexts~\cite{westhausser25longterm,10.1145/3586183.3606763}.
Existing research remains largely method-centric, focusing on personalization mechanisms and model design rather than  systematic evaluation of personalization capabilities~\cite{10.1145/3711896.3736570,hao25evaluating}.

\smallskip
\noindent\textbf{Benchmarks for Personalized Agents}
are derived from general-purpose evaluation settings ~\cite{liu2025agentbenchevaluatingllmsagents}, including QA, tool use, and short-horizon planning tasks~\cite{hu2023languagemodelsagentmodels}. These benchmarks  measure task success or accuracy under static assumptions and task-level metrics~\cite{zhu2025evolutionaryperspectivesevaluationllmbased}
. While effective for assessing general agent capabilities, they offer limited support for evaluating personalized behaviors~\cite{hao25evaluating}.
Existing personalization benchmarks operationalize personalization as profiling, either by extracting user attributes via QA or by aligning responses under explicit and static preference specifications~\cite{mok25etp,tan25personabench,zhao25personalens}.
These settings under-specify the core challenge faced by agents: synthesis, which involves  integrating implicit, potentially conflicting, and time-varying user signals into task execution over  interaction histories.
Motivated by these limitations, we introduce \textsc{IBA-Bench}. A comparison with prior benchmarks is summarized in Table~\ref{tab:persona_benchmarks}.

\begin{table}[h!]
\centering
\renewcommand{\arraystretch}{1.2}
\resizebox{\columnwidth}{!}{%
\setlength{\tabcolsep}{3.5pt}
\begin{tabular}{lccccc}
\toprule
\textbf{Benchmark} & \textbf{History} & \textbf{Dynamic} & \textbf{Implicit} & \textbf{Task} & \textbf{Behavior} \\
\midrule
HiCUPID      
& \textcolor{mygreen}{\ding{51}} 
& \textcolor{myred}{\ding{55}} 
& \textcolor{mygreen}{\ding{51}} 
& \textcolor{myred}{\ding{55}} 
& \textcolor{myred}{\ding{55}} \\

PersonaBench 
& \textcolor{mygreen}{\ding{51}} 
& \textcolor{mygreen}{\ding{51}} 
& \textcolor{mygreen}{\ding{51}} 
& \textcolor{myred}{\ding{55}} 
& \textcolor{myred}{\ding{55}} \\

PersonaLens  
& \textcolor{myred}{\ding{55}} 
& \textcolor{myred}{\ding{55}} 
& \textcolor{mygreen}{\ding{51}} 
& \textcolor{mygreen}{\ding{51}} 
& \textcolor{myred}{\ding{55}} \\

\textbf{\textsc{IBA-Bench} (Ours)} 
& \textcolor{mygreen}{\ding{51}} 
& \textcolor{mygreen}{\ding{51}} 
& \textcolor{mygreen}{\ding{51}} 
& \textcolor{mygreen}{\ding{51}} 
& \textcolor{mygreen}{\ding{51}} \\
\bottomrule
\end{tabular}%
}
\caption{Comparison of Persona-related Benchmarks. Our proposed \textsc{IBA-Bench} is the only benchmark that covers all dimensions.}
\label{tab:persona_benchmarks}

\end{table}

\section{Benchmark Construction}
\label{sec:construction}

To bridge the gap between static profiling and dynamic synthesis, we construct \textsc{IBA-Bench} through a multi-stage pipeline designed to simulate the noise, implicitness, and conflicts inherent in real-world personalization.
An \textsc{IBA-Bench} instance is defined by $(\mathcal{H}_u,x,s,\mathcal{Y},\mathcal{E})$, where $\mathcal{H}_u$ is the  interaction history of user $u$, $x$ is the current task query under scenario $s$, $\mathcal{Y}$ is the action space, and $\mathcal{E}$ is the behavioral evaluator. 
The action space $\mathcal{Y}$ includes two types:
\begin{itemize}[topsep=2pt,itemsep=1pt,parsep=0pt,leftmargin=*]
    \item \texttt{Generate(text)}: produce personalized textual outputs such as writing or planning;
    \item \texttt{CallAPI(params)}: generate structured parameters for tool-mediated actions such as ordering food or making a booking.
\end{itemize}
Given $\mathcal{H}_u$, $x$, and $s$, the agent must infer the historical signals most relevant to the current task and determine an action $y \in \mathcal{Y}$, where $y = f_\theta(\mathcal{H}_u, x, s)$.

The evaluator $\mathcal{E}$ assesses the behavioral correctness of the final action using type-specific criteria such as  generation quality and API-parameter matching. 
Detailed scoring definitions and examples are provided in Appendix~\ref{app:task_define}.
Figure~\ref{fig:fig1} shows the construction process of 
\textsc{IBA-Bench}.

\smallskip
\noindent\textbf{Construct Personas.}
We define 400 seed personas as the basic units for modeling user-specific variation. Each persona is multi-dimensional and formalized as
\( p = (B, T, E, R) \),
where \( B \) denotes background attributes, \( T \) personality traits, \( E \) transient states (e.g., stress or urgency), and \( R \) the current professional or social role. Note that \( B \) and \( T \)  are long-term attributes that remain relatively stable over time, while \( E \) and \( R \) are short-term preferences that may change and contradict earlier behavior.
This formulation allows a  persona to capture personal traits, situational states, and professional or social roles, enabling coherent yet diverse behaviors across scenarios.

\begin{table*}[t!]
\centering

\scriptsize
\setlength{\tabcolsep}{1.6pt}
\renewcommand{\arraystretch}{0.95}

\setlength{\tabcolsep}{1.8pt}
\renewcommand{\arraystretch}{0.95}%
\begin{tabular*}{\linewidth}{@{\extracolsep{\fill}} l cccc ccccc ccccc}

\toprule
\multirow{2}{*}{\textbf{Domain}} & 
\multicolumn{4}{c}{\textbf{Scale Stats}} & 
\multicolumn{5}{c}{\textbf{\# Evidence}} & 
\multicolumn{5}{c}{\textbf{\# Constraints}} \\
\cmidrule(lr){2-5} \cmidrule(lr){6-10} \cmidrule(lr){11-15}
& \textbf{S} & \textbf{D} & \textbf{P} & \textbf{I} 
& \textbf{1} & \textbf{2} & \textbf{3} & \textbf{4} & \textbf{5+}
& \textbf{1} & \textbf{2} & \textbf{3} & \textbf{4} & \textbf{5+} \\
\midrule
Writing and Design                & 16 & 55 & 10 & 750   &271 &267  &83  &61  &68  &66  &199  &282  &160  &43  \\
Work                   & 12 & 53 & 17 & 1,155 &394 &380  &165  &111  &105  &30  &161  &338  &321  &305 \\
Daily Consumption      & 8  & 39 & 15 & 1,086 &511 &290  &217  &36  &32  &46  &182  &401  &276  &181 \\
Planning               & 10 & 39 & 16 & 1,151 &532 &329  &182  &42  &66  &41  &206  &489  &386  &29  \\
Exercise and Health    & 4  & 16 & 20 & 781   &325 &247  &113  &41  &55  &11  &54   &173  &252  &291 \\
Transportation         & 4  & 14 & 14 & 518   &288 &117  &112  &0   &1   &21  &67   &169  &197  &64  \\
Medical Services       & 3  & 11 & 15 & 524   &290 &113  &121  &0   &0   &18  &90   &231  &185  &0   \\
Leisure Activities     & 3  & 9  & 14 & 333   &137 &104  &62   &16  &14  &12  &60   &148  &113  &0   \\
Information Management & 6  & 26 & 17 & 664   &293 &217  &94   &29  &31  &23  &73   &185  &161  &222 \\
\midrule
\textbf{Overall}       & \textbf{66} & \textbf{262} & 
 \textbf{15} & \textbf{6,962} &\textbf{3,041}  &\textbf{2,064}  &\textbf{1,149} &\textbf{336} &\textbf{372}  &\textbf{268}  &\textbf{1,092}  &\textbf{2,416}  &\textbf{2,051}  &\textbf{1,135} \\
\bottomrule
\end{tabular*}

\caption{
Statistics of domains in the benchmark.
\textbf{S}: number of scenarios in each domain;
\textbf{D}: total number of preference dimensions across scenarios;
\textbf{P}: integer-truncated average number of preference values per scenario;
\textbf{I}: number of task instances.
The columns under \textbf{\#Evidence} and \textbf{\#Constraints} report the distributions of task instances by the number of associated evidence items and preference constraints in each instance, respectively.
In both column groups, ``5+'' indicates five or more.
For each domain, the counts under \textbf{\#Evidence} and \textbf{\#Constraints} each sum to \textbf{I}.
}
\label{tab:domain_summary}
\end{table*}

\smallskip
\noindent\textbf{Generate Interaction History.}
To mimic realistic interaction history, we use a multi-agent framework to generate 
interactions where informative preference evidence is embedded within large volumes of noisy conversational content. Rather than being explicitly stated, user preferences are revealed  through accumulated behavioral patterns. For example, instead of directly stating a preference such as “I prefer concise reports,” a user may repeatedly shorten verbose drafts or prioritize rapid responses under time pressure. Consequently, agents must infer preferences from behavioral evidence from the interaction history.

To increase realism, we inject   task-irrelevant interactions, with task-relevant interactions occurring only once per 100 turns on average, causing informative signals to be deeply buried within noisy histories. This signal-to-noise imbalance challenges agents to distinguish meaningful behavioral evidence from distractors.

\begin{figure*}[t]
    \centering
    \includegraphics[width=\textwidth]{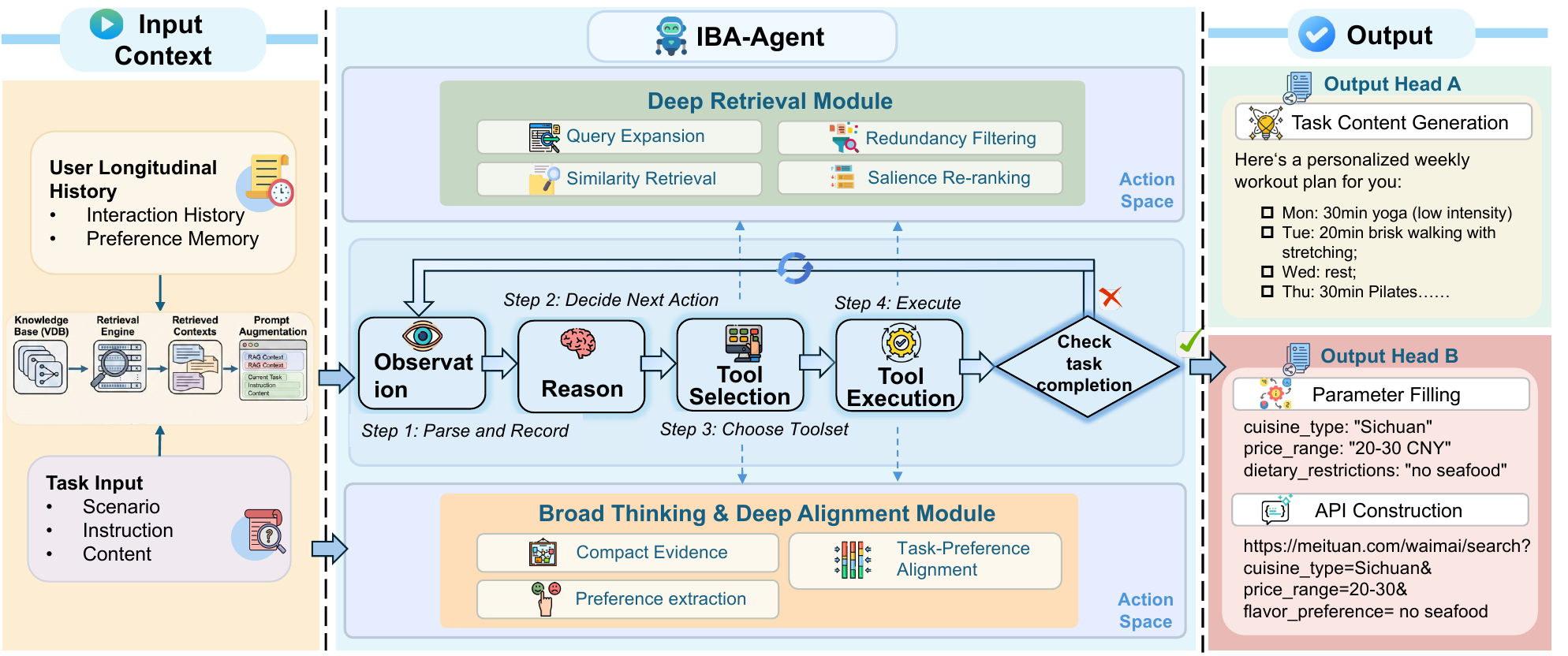}
    \caption{Overview of IBA-Agent for personalized task completion.}
    \vspace{-4mm}
    \label{fig:pic5}
\end{figure*}

\smallskip
\noindent\textbf{Task Instantiation and Validation.}
\footnotetext[1]{\url{https://huggingface.co/BAAI/bge-m3}}
We use  minimal  prompts to generate  tasks that require agents to execute concrete actions based on the interaction history. Task content  is newly  authored but domain-aligned to prevent trivial reuse. 
To 
simulate real-world agent operations, we include  2,280 API-actionable instances spanning 16 scenarios with parameterized API-call outputs.

Table~\ref{tab:domain_summary} shows the statistics of  \textsc{IBA-Bench} which comprises 6,962 task instances 
spanning 9 domains across 66 scenarios and 262 preference dimensions, generated using GPT-5.1 at an API cost of approximately \$5,000.
The distribution is skewed toward non-trivial cases: 1,857 instances involve three or more behavioral evidence items, 5,602 instances contain at least three preference constraints
(see Appendix~\ref{app:dataset_details}).

For quality control, we randomly sample 200 instances for validation. 
Across reasonableness, coherence, and persona consistency, human annotations achieve 80.2\% agreement with a Gwet's $AC_1$ of 0.70, while AI-based annotations achieve 94.0\% agreement with an $AC_1$ of 0.93.
Based on the validation feedback, we  refine the prompts and regenerate problematic cases,  revising about  10\% of all instances. Annotation details are in Appendix~\ref{app:human_validation}.

\section{Method}
\label{sec:method}
In this section, we formalize implicit behavioral alignment as a preference-conditioned task execution problem and present IBA-Agent, an LLM-driven personalized agent framework for bridging the gap between historical preference understanding and active task execution, as shown in Figure~\ref{fig:pic5}.

\subsection{LLM-Driven Personalized Agent}
\label{subsec:agent}
IBA-Agent takes the current task and the user's historical interactions as input, and follows an iterative evidence-to-action process for personalized task execution. 
During this process, an LLM controller maintains the task state, observes the current context, reasons over accumulated tool outputs, selects and executes the next tool, and updates the context with new evidence or intermediate decisions. 
Through this loop, the agent identifies dynamic and implicit preference signals from the user's history and aligns them with the requirements of the current task. 
The process terminates when the controller invokes a task-specific final output head to produce the personalized action.
IBA-Agent implements this workflow with two complementary modules: \textsc{Deep Retrieval} and \textsc{Broad Thinking \& Deep Alignment}, which we detail below.

\smallskip
\noindent\textbf{Deep Retrieval.   }
This module uses four retrieval tools to identify task-relevant preference evidence from the large interaction history $\mathcal{H}_u$ and dynamically build the task-conditioned evidence context $\mathcal{C}_u(x)$ for downstream reasoning.
Given a current task query $x$, the agent first retrieves passages using the original task description and then invokes \texttt{query expansion} to generate multiple task-aware sub-queries. As shown in Table~\ref{tab:query_expansion_prompt}, these sub-queries probe complementary preference facets.
This design allows the same user history to yield task-specific evidence, such as budget trade-offs for purchase tasks and tone, formatting, and revision patterns for writing tasks.
The \texttt{similarity retrieval} tool, implemented with BGE-M3, retrieves semantically relevant passages from $\mathcal{H}_u$.
This enables the agent to recover implicit preference evidence even when the user has not stated a preference in a canonical form.

\begin{table}[t]
\centering

\definecolor{softtitle}{RGB}{92,92,92}
\definecolor{softblue}{RGB}{38,70,145}
\definecolor{softgreen}{RGB}{42,145,62}
\definecolor{softred}{RGB}{185,55,55}
\definecolor{softpurple}{RGB}{170,75,220}
\definecolor{softgray}{RGB}{95,95,95}

\begin{tcolorbox}[
    enhanced,
    width=0.98\columnwidth,
    colback=gray!2,
    colframe=gray!50,
    boxrule=0.7pt,
    arc=2pt,
    left=6pt,
    right=6pt,
    top=5pt,
    bottom=5pt,
    drop shadow={black!10!white},
    title={\textsc{Query Expansion Prompt}},
    coltitle=white,
    colbacktitle=softtitle,
    fonttitle=\bfseries,
    fontupper=\ttfamily\scriptsize,
    boxed title style={
        sharp corners,
        arc=1pt,
        boxrule=0pt,
        left=6pt,
        right=6pt,
        top=2pt,
        bottom=2pt,
        colback=softtitle
    }
]

\setlength{\baselineskip}{1.02\baselineskip}

\noindent
You are a \textbf{\textcolor{softblue}{search query generator}} for personalized recommendations. 
Given a task query, generate diverse retrieval queries to uncover relevant user preferences from historical conversations.

\vspace{0.3\baselineskip}

\raggedright
\noindent
\textbf{\textcolor{softblue}{Input:}} 
Task query information, 
including the scenario, instruction, and content.

\vspace{0.35\baselineskip}

\noindent
\textbf{\textcolor{softgreen}{Expansion Facets.}}
\vspace{0.1\baselineskip}

\begin{tabularx}{\linewidth}{p{0.23\linewidth} X}
\toprule
\textbf{Facet} & \textbf{Retrieval focus} \\
\midrule

$\textcolor{softpurple}{q^{style}}$ 
& Tone, wording, expression habits, and preferred writing style. \\
\arrayrulecolor{gray!45}\midrule\arrayrulecolor{black}

$\textcolor{softpurple}{q^{structure}}$ 
& Format, headings, templates, bullet/paragraph preference, length, and detail level. \\
\arrayrulecolor{gray!45}\midrule\arrayrulecolor{black}

$\textcolor{softpurple}{q^{decision}}$ 
& Trade-offs such as cost vs. quality, time vs. thoroughness, or convenience vs. preference. \\
\arrayrulecolor{gray!45}\midrule\arrayrulecolor{black}

$\textcolor{softpurple}{q^{correction}}$ 
& Complaints, corrections, explicit rejections, and disliked suggestions. \\
\arrayrulecolor{gray!45}\midrule\arrayrulecolor{black}

$\textcolor{softpurple}{q^{habit}}$ 
& Repeated choices, default behaviors, and recurring preferences in the scenario. \\

\bottomrule
\end{tabularx}

\vspace{0.35\baselineskip}

\noindent
\textbf{\textcolor{softred}{Output:}} 
Return target query strings covering multiple facets. 
No preamble or explanation.

\end{tcolorbox}

\caption{Prompt-guided query expansion in the Deep Retrieval module. Given a task $x$, the generator produces a set of target queries $Q(x)=\{q^{style}, q^{structure}, q^{decision}, q^{correction}, q^{habit}\}$ to retrieve diverse preference evidence from the history.}
\label{tab:query_expansion_prompt}
\end{table}

\begin{table*}[!t]
\centering
\footnotesize
\setlength{\tabcolsep}{5.5pt}
\renewcommand{\arraystretch}{1.08}
\resizebox{1.0\linewidth}{!}{
\begin{tabular}{l *{10}{c}}
\toprule
\textbf{Model} & \textbf{Writing} & \textbf{Work} & \textbf{Daily} & \textbf{Plan.} & \textbf{Health} & \textbf{Trans.} & \textbf{Med.} & \textbf{Leis.} & \textbf{Info.} & \textbf{Overall} \\
\midrule
\specialrule{0.08em}{0.15em}{0.15em}
\multicolumn{11}{c}{\textbf{\textit{$\bullet$ Baseline Models}}} \\
\specialrule{0.08em}{0.15em}{0.15em}
Qwen2.5-7B-Instruct~\cite{qwen2025qwen25technicalreport} & 52.7 & 51.0 & 77.3 & 55.5 & 67.1 & 65.2 & 75.0 & 40.1 & 37.1 & 58.3 \\
DeepSeek-V3~\cite{deepseekai2025deepseekv3technicalreport} & 65.1 & 56.6 & 77.1 & 62.7 & 71.4 & 67.3 & 78.8 & 44.7 & 51.9 & 64.9 \\
ChatGLM-4-9b-chat~\cite{glm2024chatglmfamilylargelanguage} & 51.4 & 47.3 & 70.7 & 53.5 & 65.4 & 58.4 & 69.3 & 36.7 & 37.1 & 55.3 \\

QwQ-32B~\cite{QwQ32BBlog2025} & 68.7 & 61.6 & 80.6 & 64.5 & 75.2 & 74.2 & 79.6 & 47.3 & 55.4 & 68.4 \\
GPT-4o-mini & 57.7 & 54.9 & 77.4 & 57.9 & 68.1 & 68.3 & 71.8 & 39.9 & 42.4 & 61.1 \\
GPT-4o~\cite{OpenAI_GPT4o2024} & 63.5 & 54.1 & 75.6 & 57.9 & 69.5 & 66.6 & 73.0 & 44.1 & 43.5 & 61.7 \\
GPT-5-mini & 70.5 & 63.9 & 76.3 & 69.8 & 78.3 & 70.8 & 80.2 & 49.8 & 63.2 & 70.1 \\
GPT-5.1~\cite{OpenAI_GPT5_1_2025} & 72.9 & 62.4 & 79.9 & 65.6 & 73.9 & 70.6 & 78.0 & \best{56.8} & 58.1 & 69.2 \\

\midrule
\specialrule{0.08em}{0.15em}{0.15em}
\multicolumn{11}{c}{\textbf{\textit{$\bullet$ Qwen3-4B-Instruct Variants}}} \\
\specialrule{0.08em}{0.15em}{0.15em}
Qwen3-4B-Instruct~\cite{qwen3technicalreport} & 65.2 & 60.4 & 77.8 & 62.8 & 73.4 & 80.0 & 82.9 & 52.6 & 54.1 & 67.6 \\
IBA-Agent (Qwen3-4B-Instruct) & 80.4 & 73.3 & 84.0 & 77.2 & 86.1 & \best{82.3} & 89.5 & 47.5 & 69.2 & 78.1 \\
\rowcolor{lightpurple}
IBA-Agent (Qwen3-4B-Instruct, FC) & 80.8 & 73.9 & 83.0 & \uline{79.3} & \uline{87.7} & \uline{81.9} & \uline{91.4} & 44.9 & 69.8 & 78.7 \\

\midrule
\specialrule{0.08em}{0.15em}{0.15em}
\multicolumn{11}{c}{\textbf{\textit{$\bullet$ DeepSeek-V3.2 Variants}}} \\
\specialrule{0.08em}{0.15em}{0.15em}
DeepSeek-V3.2~\cite{deepseekai2025deepseekv32pushingfrontieropen} & 68.1 & 60.1 & 79.8 & 63.2 & 72.2 & 70.7 & 79.0 & 48.3 & 53.6 & 66.9 \\
IBA-Agent (DeepSeek-V3.2) & \uline{83.5} & \uline{75.1} & \uline{84.5} & 76.9 & 85.6 & 78.0 & 87.9 & 53.3 & 72.3 & \uline{78.8} \\
\rowcolor{lightpurple}
IBA-Agent (DeepSeek-V3.2, FC) & \best{86.7} & \best{77.7} & \best{88.9} & \best{79.8} & \best{89.7} & 80.2 & \best{92.6} & \uline{54.2} & \best{74.1} & \best{81.9} \\

\midrule
\specialrule{0.08em}{0.15em}{0.15em}
\multicolumn{11}{c}{\textbf{\textit{$\bullet$ General-Purpose AI Agents}}} \\
\specialrule{0.08em}{0.15em}{0.15em}

nanobot (DeepSeek-V3.2) ~\cite{ren25nanobot} & 69.9 & 54.0 & 59.3 & 62.5 & 56.2 & 53.0 & 59.7 & 38.1 & 61.1 & 58.5 \\
Claude Code (DeepSeek-V3.2) \footnotemark[4]& 80.5 & 64.0 & 60.9 & 70.3 & 63.1 & 54.7 & 58.1 & 33.8 & 56.7 & 62.9 \\
Hermes Agent(DeepSeek-V3.2) ~\cite{nous25hermes} & 81.7 & 63.8 & 54.2 & 70.9 & 59.0 & 56.9 & 59.5 & 35.9 & 68.6 & 63.1 \\

\bottomrule
\end{tabular}
}
\caption{Performance comparison using the \textbf{bge-m3} retriever. Results are evaluated by DeepSeek-V3.2, Qwen3-4B-Instruct, and Qwen3-30B-A3B-Instruct; scores from Qwen3-4B-Instruct are averaged over five runs. Columns correspond to 9 application domains: \textbf{Writing} (writing and design), \textbf{Work}, \textbf{Daily} (daily consumption), \textbf{Plan.} (planning), \textbf{Health} (exercise and health), \textbf{Trans.} (transportation), \textbf{Med.} (medical services), \textbf{Leis.} (leisure activities), and \textbf{Info.} (information management). Best result is highlighted in \colorbox{bestgreen}{\textbf{green bold text}},  second-best result is \uline{underlined}, \colorbox{lightpurple}{Function Call (FC) variants} are shaded in light purple.}
\label{tab:res_bge}
\vspace{-3mm}
\end{table*}

The module also provides evidence refinement tools. The \texttt{redundancy filtering} tool removes duplicated or highly overlapping retrieved passages using a content-hash heuristic, preventing repeated history snippets from dominating the evidence context. 
The \texttt{salience re-ranking} tool uses an LLM-based scorer to prioritize passages with high-confidence preference signals. 
Each passage is assigned a discrete salience score from 0 to 10, and passages scoring at least 5 are retained for downstream use; the scoring criteria are summarized in Table~\ref{tab:salience_score} in the Appendix. 
This weighting scheme limits over-interpretation of weak behavioral cues by weighting reliable explicit feedback more heavily than noisier implicit behavioral patterns.

\noindent\textbf{Broad Thinking \& Deep Alignment.   }
This module uses a set of prompt-guided reasoning tools to transform the $\mathcal{C}_u(x)$ into a task-specific personalization plan. The \texttt{compact evidence} function first organizes the retained passages into a concise evidence set by preserving the most informative historical interactions. The \texttt{preference extraction} function then reads this evidence with respect to the current task and extracts task-relevant preference signals, such as what the user asked for, corrected, rejected, repeatedly chose, or recently updated. The \texttt{task-preference alignment} function decides how the extracted preferences should guide the current task, producing an alignment plan: response style and structure for generation tasks and constraints or priorities for API-action tasks. The synthesis step then realizes this plan as the final output, namely personalized content or filled API parameters with an action description.

\begin{figure}[t]
    \centering
    \includegraphics[
        width=0.88\columnwidth,
        trim=2mm 5mm 2mm 4mm,
        clip
    ]{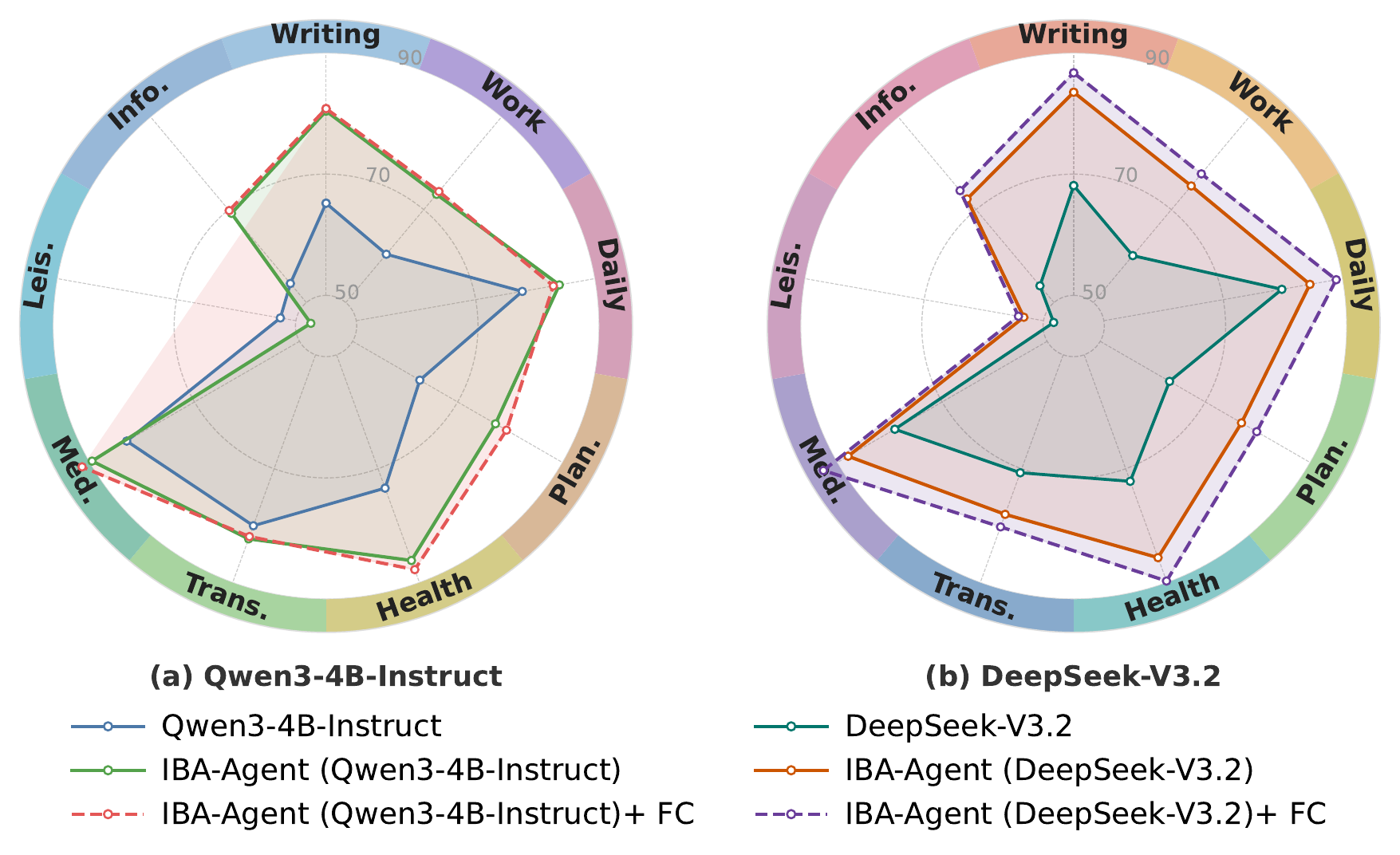}
    \caption{Radar plots of baseline models and IBA-Agent variants across nine domains.}
    \label{fig:pic6}
    \vspace{-2mm}
\end{figure}

\begin{table*}[!ht]
    \centering
    \small
    \setlength{\tabcolsep}{6pt}
    \resizebox{1.01\linewidth}{!}{
    \begin{tabular}{l cccccccccc}
        \toprule
        \textbf{Method Variants}
        & \textbf{Writing} & \textbf{Work} & \textbf{Daily} & \textbf{Plan.}
        & \textbf{Health} & \textbf{Trans.} & \textbf{Medical}
        & \textbf{Leisure} & \textbf{Info.} & \textbf{Overall} \\
        \midrule

        \rowcolor{lightpurple}

\textbf{IBA-Agent (DeepSeek-V3.2) }
& \textbf{83.5} & \textbf{75.1} & \textbf{84.5} & \textbf{76.9}
& \textbf{85.6} & \textbf{78.0} & \textbf{87.9}
& \textbf{53.3} & \textbf{72.3} & \textbf{78.8} \\
\quad w/o Deep Retrieval
& 79.7{\scriptsize\textcolor{myred}{$\downarrow$3.8}}
& 67.4{\scriptsize\textcolor{myred}{$\downarrow$7.7}}
& 72.9{\scriptsize\textcolor{myred}{$\downarrow$11.6}}
& 73.8{\scriptsize\textcolor{myred}{$\downarrow$3.1}}
& 79.3{\scriptsize\textcolor{myred}{$\downarrow$6.3}}
& 65.1{\scriptsize\textcolor{myred}{$\downarrow$12.9}}
& 64.9{\scriptsize\textcolor{myred}{$\downarrow$23.0}}
& 46.7{\scriptsize\textcolor{myred}{$\downarrow$6.6}}
& 65.9{\scriptsize\textcolor{myred}{$\downarrow$6.4}}
& 70.5{\scriptsize\textcolor{myred}{$\downarrow$8.3}} \\

\quad w/o BT+DA
& 82.7{\scriptsize\textcolor{myred}{$\downarrow$0.8}}
& 71.9{\scriptsize\textcolor{myred}{$\downarrow$3.2}}
& 83.8{\scriptsize\textcolor{myred}{$\downarrow$0.7}}
& 77.4{\scriptsize\textcolor{mygreen}{$\uparrow$0.5}}
& 84.4{\scriptsize\textcolor{myred}{$\downarrow$1.2}}
& 74.1{\scriptsize\textcolor{myred}{$\downarrow$3.9}}
& 82.7{\scriptsize\textcolor{myred}{$\downarrow$5.2}}
& 48.4{\scriptsize\textcolor{myred}{$\downarrow$4.9}}
& 70.6{\scriptsize\textcolor{myred}{$\downarrow$1.7}}
& 76.3{\scriptsize\textcolor{myred}{$\downarrow$2.5}} \\

\quad w/o Synth.
& 76.4{\scriptsize\textcolor{myred}{$\downarrow$7.1}}
& 67.0{\scriptsize\textcolor{myred}{$\downarrow$8.1}}
& 83.1{\scriptsize\textcolor{myred}{$\downarrow$1.4}}
& 70.6{\scriptsize\textcolor{myred}{$\downarrow$6.3}}
& 81.6{\scriptsize\textcolor{myred}{$\downarrow$4.0}}
& 75.8{\scriptsize\textcolor{myred}{$\downarrow$2.2}}
& 86.3{\scriptsize\textcolor{myred}{$\downarrow$1.6}}
& 48.1{\scriptsize\textcolor{myred}{$\downarrow$5.2}}
& 60.6{\scriptsize\textcolor{myred}{$\downarrow$11.7}}
& 73.3{\scriptsize\textcolor{myred}{$\downarrow$5.5}} \\
        \bottomrule
    \end{tabular}
    }
    \caption{
    Ablation study of IBA-Agent with DeepSeek-V3.2 as the backbone LLM across task domains, using \textbf{bge-m3} as the retriever. Results are evaluated by three judges: DeepSeek-V3.2, Qwen3-4B-Instruct, and Qwen3-30B-A3B-Instruct; Qwen3-4B-Instruct scores are averaged over five runs. 
    \textit{w/o} denotes removing a specific component.
    \textbf{BT+DA} denotes \textit{Broad Thinking + Deep Alignment}, and
    \textbf{Synth.} denotes the original RAG-style synthesis strategy without trajectory-level synthesis.
    }
    \vspace{-3mm}
    \label{tab:ablation}
\end{table*}

\vspace{-1mm}

\begin{figure*}[t]
    \centering
    \setlength{\tabcolsep}{6pt}
    \renewcommand{\arraystretch}{1.0}

    \begin{tabular}{@{\hspace*{-6mm}}c c@{}}
        \begin{subfigure}[t]{0.47\textwidth}
            \centering



        


\begin{tikzpicture}
\begin{axis}[
    xshift=-15mm,  
    ybar,
    bar width=8pt,
    width=1.05\columnwidth,      
    height=4.1cm,
    ylabel={Score (\%)},
    ymin=50, ymax=80,
    xmin=0.5, xmax=3.5,
    xtick={1,2,3},
    axis line style={black!55},
    tick style={black!55},
    bar shift auto,
    tick label style={font=\scriptsize},
    label style={font=\small},
    ymajorgrids=true,
    grid style=dashed,
    clip=false,                  
    legend cell align=left,
    legend style={
        at={(0.5,1.0)},          
        anchor=south,
        legend columns=2,         
        draw=none,
        fill=none,                
        font=\scriptsize,
        column sep=3pt,
        row sep=-1pt,
    },
]
\addplot[fill=mygreen!35, draw=mygreen, line width=0.4pt]
    coordinates {(1,58.8) (2,59.1) (3,56.9)};
\addplot[fill=myred!30, draw=myred, line width=0.4pt]
    coordinates {(1,58.5) (2,61.2) (3,53.7)};
\addplot[fill=myblue!35,  draw=myblue,  line width=0.4pt]
    coordinates {(1,63.3) (2,65.6) (3,55.9)};

\addplot[fill=myorange!45, draw=myorange, line width=0.4pt]
    coordinates {(1,65.8) (2,72.0) (3,64.7)};
\addplot[fill=mypurple!45, draw=mypurple, line width=0.4pt]
    coordinates {(1,73.1) (2,77.7) (3,73.1)};

\legend{
Standard RAG (QwQ-32B),
Standard RAG (GPT-5.1),
Standard RAG (GPT-4o),
IBA-Agent (w/o Deep Retrieval),
IBA-Agent
}

\node[font=\scriptsize, anchor=south] at (axis cs:1,74.1) {$\Delta{=}7.3$};
\node[font=\scriptsize, anchor=south] at (axis cs:2,75.9) {$\Delta{=}5.7$};
\node[font=\scriptsize, anchor=south] at (axis cs:3,74.1) {$\Delta{=}8.4$};

\end{axis}
\end{tikzpicture}
            \vspace{-6mm}
            \caption{Impact of number of evidence.
            }
            \label{fig:multi_source}
        \end{subfigure}
        &
        \begin{subfigure}[t]{0.47\textwidth}
            \centering
            \input{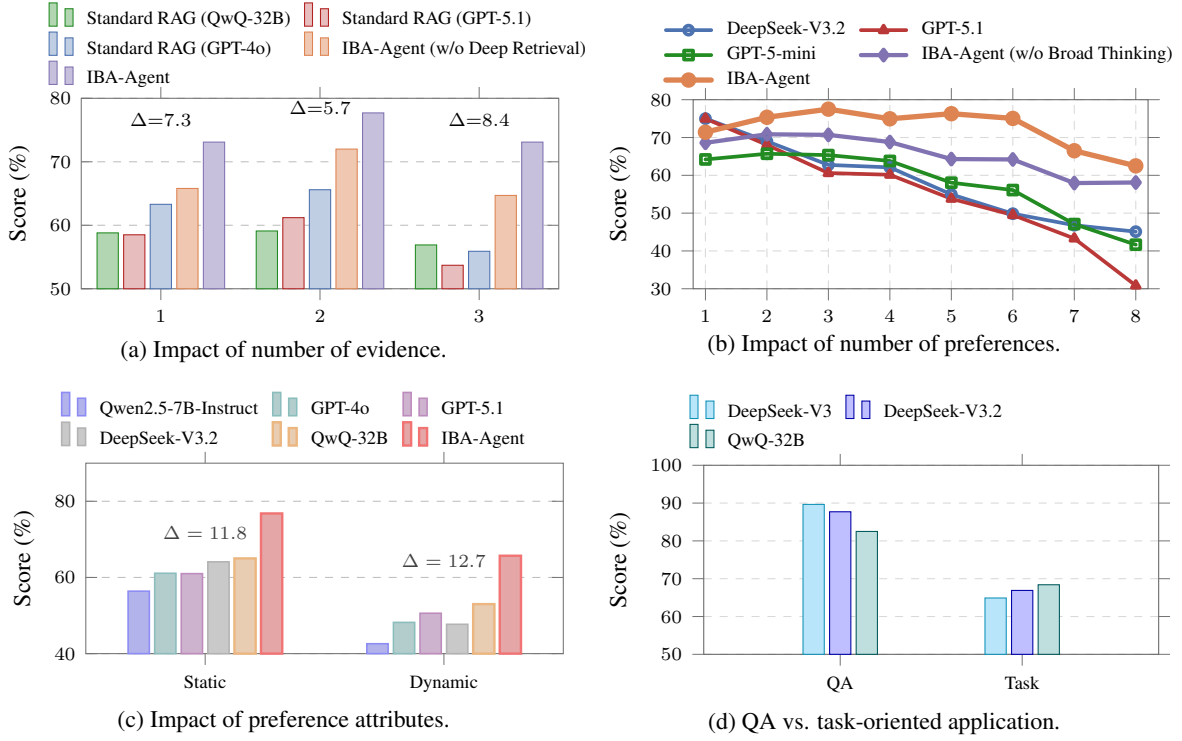}
            \vspace{-2mm}
            \caption{Impact of number of preferences.}
            \label{fig:quantity}
        \end{subfigure}
        \\
        \vspace{-1mm} & \vspace{-1mm} \\
        \begin{subfigure}[t]{0.47\textwidth}
            \centering
\begin{tikzpicture}
\begin{axis}[
    ybar,
    bar width=8pt,
    width=1.05\columnwidth,
    height=4.1cm,
    ylabel={Score (\%)},
    ymin=40, ymax=90,
    xmin=0.5, xmax=2.5,
    xtick={1,2},
    xticklabels={Static, Dynamic},
    bar shift auto,
    tick label style={font=\scriptsize},
    label style={font=\small},
    axis line style={black!55},
    tick style={black!55},
    ymajorgrids=true,
    grid style=dashed,
    clip=false, 
    legend style={
        at={(-0.07,1.02)},
        anchor=south west,
        legend columns=3,
        draw=none,
        fill=none,
        font=\scriptsize,
        column sep=3pt,
        row sep=-1pt,
        cells={anchor=west},
    },
    legend cell align=left,
]
\addplot[
  fill=blue!75!black!30,
  draw=blue!90!black!45,
  line width=0.6pt
]
coordinates {(1,56.4) (2,42.6)};

\addplot[
  fill=teal!70!black!30,
  draw=teal!85!black!45,
  line width=0.6pt
]   coordinates {(1,61.1) (2,48.2)};

\addplot[
  fill=violet!70!black!30,
  draw=violet!85!black!45,
  line width=0.6pt
]    coordinates {(1,61.0) (2,50.6)};

\addplot[
  fill=gray!60!black!30,
  draw=gray!80!black!50,
  line width=0.6pt
]
    coordinates {(1,64.1) (2,47.7)};

\addplot[
  fill=orange!70!black!30,
  draw=orange!85!black!50,
  line width=0.9pt
]
    coordinates {(1,65.0) (2,53.0)};

\addplot[
  fill=red!70!black!35,
  draw=red!90!black!55,
  line width=0.9pt
]    coordinates {(1,76.8) (2,65.7)};

\node[font=\scriptsize\bfseries, anchor=south, text=black!75] at (axis cs:1,67.5) {$\Delta = 11.8$};
\node[font=\scriptsize\bfseries, anchor=south, text=black!75] at (axis cs:2,60) {$\Delta = 12.7$};

\legend{Qwen2.5-7B-Instruct, GPT-4o, GPT-5.1, DeepSeek-V3.2, QwQ-32B, IBA-Agent}
\end{axis}
\end{tikzpicture}
            \vspace{-6mm}
            \caption{Impact of preference attributes.}
            \label{fig:dynamic}
        \end{subfigure}
        &
        \begin{subfigure}[t]{0.47\textwidth}
            \centering
            \begin{tikzpicture}
\begin{axis}[
    ybar,
    scale only axis,
    bar width=8pt,
    width=0.8\columnwidth,
    height=2.5cm,
    xmin=0.75,
    xmax=2.25,
    xtick={1.20,1.80},
    xticklabels={QA, Task},
    xticklabel style={font=\scriptsize},
    ylabel={Score (\%)},
    yticklabel style={font=\scriptsize},
    ylabel style={font=\small},
    ymin=50,
    ymax=100,
    ytick={50,60,70,80,90,100},
    tick align=outside,
    tick style={black!55},
    axis line style={black!60},
    ymajorgrids=true,
    grid style={dashed, black!15},
    legend style={
        at={(-0.05,1.02)},
        anchor=south west,
        legend columns=2,
        draw=none,
        fill=none,
        font=\scriptsize,
        column sep=3pt,
        row sep=-1pt,
        cells={anchor=west},
    },
    legend cell align=left,
    legend image post style={xscale=1.0, yscale=1.0},
]

\addplot+[draw=cyan!70!black, fill=cyan!25, bar shift=-10pt]
coordinates {(1.20,89.65) (1.80,64.9)};
\addlegendentry{DeepSeek-V3}

\addplot+[draw=blue!65!black, fill=blue!25, bar shift=0pt]
coordinates {(1.20,87.7) (1.80,66.9)};
\addlegendentry{DeepSeek-V3.2}

\addplot+[draw=teal!70!black, fill=teal!25, bar shift=10pt]
coordinates {(1.20,82.5) (1.80,68.4)};
\addlegendentry{QwQ-32B}

\end{axis}
\end{tikzpicture}
            \vspace{-1mm}
            \caption{QA vs. task-oriented application.}
            \label{fig:know_gap}
        \end{subfigure}
    \end{tabular}

    \caption{Ablation analyses on evidence quantity, preference complexity, preference attributes, and the knowledge-to-action gap. $\Delta$ denotes the performance gap between IBA-Agent and the second-best model.
    }
    \label{fig:four_grid}
    \vspace{-3mm}
\end{figure*}

\section{Performance Study}

All experiments are conducted in an inference-only setting, with no parameter fine-tuning.
We evaluate \textbf{ten modern LLMs} spanning the Qwen, DeepSeek, GPT, ChatGLM, and QwQ families, together with \textbf{three general-purpose agent systems} (Claude Code~\footnotemark[4], Hermes Agent~\cite{nous25hermes}, and nanobot~\cite{ren25nanobot}), under a unified RAG pipeline that uses \textbf{bge-m3} for dense retrieval.
Each output is scored against scenario-specific \emph{behavioral checkpoints} by an LLM judge; for robustness, we aggregate results over \textbf{three judges} (DeepSeek-V3.2, Qwen3-4B-Instruct, and Qwen3-30B-A3B-Instruct).
Details of baselines, evaluation metrics, and hyperparameters are provided in Appendix~\ref{sec:exp_settings}, with additional experiments on retriever and judge robustness reported in Appendix~\ref{sec:supp_experiments}.

\footnotetext[4]{\url{https://github.com/oboard/claude-code-rev}}

\begin{figure*}[t]
    \centering
    \includegraphics[width=\textwidth]{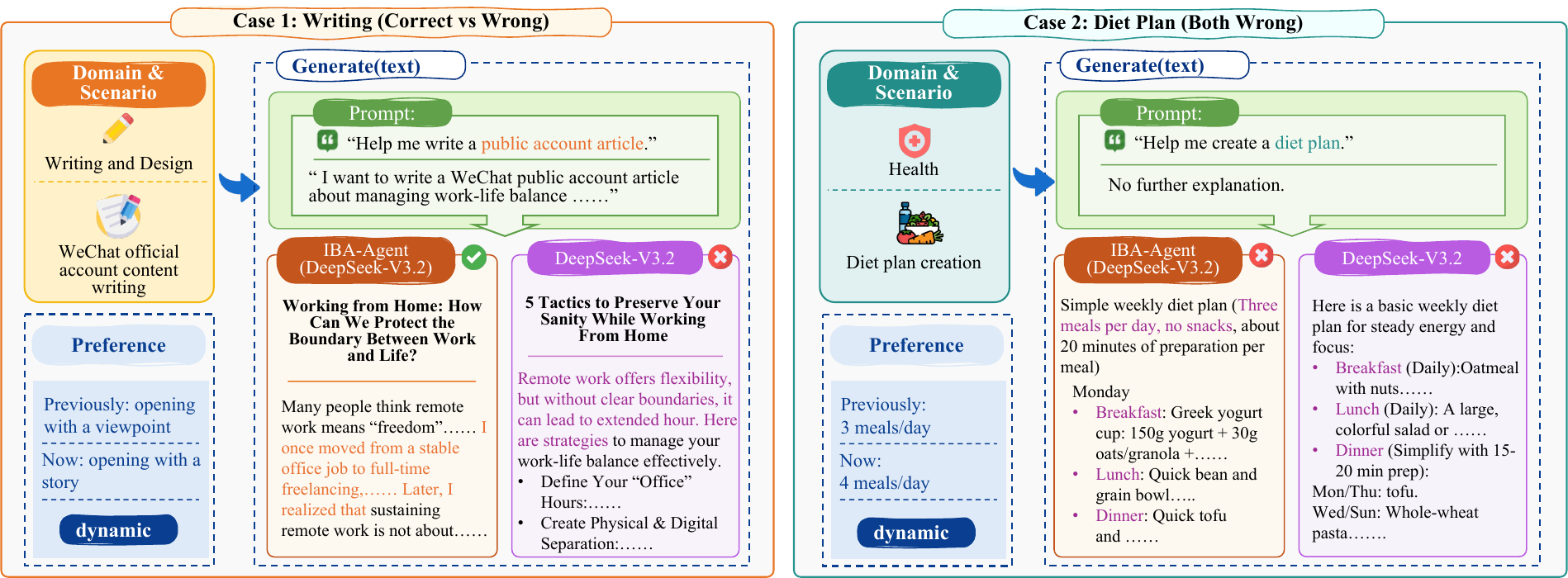}
    \caption{Case studies of personalized \texttt{Generate(text)} execution in \textsc{IBA-Bench}. 
Each case compares \textsc{IBA-Agent (Function Call)} instantiated with DeepSeek-V3.2 against the same DeepSeek-V3.2 baseline on the same task and with the same dynamic-preference evidence.}
    \label{fig:case_study_1}
    \vspace{-3mm}
\end{figure*}

\subsection{Main Results}
Table~\ref{tab:res_bge} reports results on \textsc{IBA-Bench}. Standard RAG yields moderate and inconsistent results, with gains not strictly monotonic with model scale, suggesting that
stronger backbones alone are insufficient for
personalized task completion. In contrast, IBA-Agent consistently improves over its RAG backbones, increasing DeepSeek-V3.2 from 66.9 to 78.8 and Qwen3-4B-Instruct from 67.6 to 78.1. Adding function calling further boosts performance to 81.9 and 78.7 respectively. This indicates that the key challenge lies not only in retrieving relevant memories, but also in converting implicit user preferences into actionable constraints. 

Compared to the general-purpose agent systems Claude Code, Hermes Agent, and nanobot, IBA-Agent performs substantially better, demonstrating that generic agent workflows transfer poorly to personalized settings. As shown in Figure~\ref{fig:pic6}, IBA-Agent achieves broad gains across most task categories, while function calling enables more flexible module invocation and more reliable personalized task completion.

\subsection{Ablation Study}

Table~\ref{tab:ablation} presents the ablation results of IBA-Agent on a 1,000-instance test set obtained through stratified sampling from \textsc{IBA-Bench}. The full model consists of Deep Retrieval, Broad Thinking \& Deep Alignment, and trajectory-level synthesis. To assess the contribution of each component, we consider three variants: removing Deep Retrieval, removing Broad Thinking \& Deep Alignment, and replacing trajectory-level synthesis with the original RAG-style synthesis.

Among these components, Deep Retrieval has the largest impact: removing it reduces overall performance from 78.8 to 70.5 (-8.3), with the largest drops observed in \textit{Medical}, \textit{Transportation}, and \textit{Daily}. This highlights the importance of 
accurately accessing relevant evidence from long interaction histories. Removing Broad Thinking \& Deep Alignment lowers performance from 78.8 to 76.3 (-2.5). Replacing trajectory-level synthesis with the original RAG-style alternative decreases performance to 73.3 (-5.5), suggesting that effective personalization requires not only retrieving relevant evidence but also integrating it into coherent behavioral trajectories.
Notably, these results suggest that the modules are effective as a coordinated mechanism: without reliable Deep Retrieval evidence, downstream Broad Thinking \& Deep Alignment may become counterproductive.

\vspace{-1mm}
\subsection{In-depth Analysis}

\noindent\textbf{Impact of Number of Evidence.} Figure~\ref{fig:multi_source} reports preference-conditioned task performance when the number of evidence varies. IBA-Agent consistently outperforms the standard RAG baseline and the variant without Deep Retrieval,
highlighting its ability to aggregate  preference-relevant signals and the value of broad, deep retrieval. Across all models,  increasing evidence from one to two helps, but expanding to three often hurts, suggesting a trade-off between recovering missing preferences and accumulating noise. 

\noindent\textbf{Impact of Number of Preferences.}
Figure~\ref{fig:quantity} shows the performance as  the number of preferences (smoothed with a sliding window, $w{=}2$) increases. Performance drops for all models as constraints accumulate. IBA-Agent degrades more slowly and widens its lead as  preference increases. Removing Broad Thinking \& Deep Alignment causes a clear drop, confirming its role in robust multi-constraint alignment.

\noindent\textbf{Impact of Preference Attributes.}
Figure~\ref{fig:dynamic} shows the effect of static vs. dynamic preference settings. All models perform better on static preferences, indicating easier behavior alignment.
For dynamic preferences, 
IBA-Agent outperforms the strongest baseline by +12.7 points (vs. +11.8 static), demonstrating robustness to preference shifts.

\noindent\textbf{QA vs. Task-oriented Application.}
Figure~\ref{fig:know_gap} compares QA with direct task execution. For all models, QA consistently exceeds Task, exposing a clear knowledge-to-action gap: models can state preferences correctly but often fail to enforce them, showing that QA competence does not translate into robust preference-aware task completion. 
A qualitative illustration of this QA-to-execution distinction is provided in Figure~\ref{fig:case_study_2} in the Appendix.

\subsection{Case Study on Knowledge-to-Action Gap   }
Figure~\ref{fig:case_study_1} shows two \texttt{Generate(text)} cases under dynamic preferences. 
Both cases use DeepSeek-V3.2 as the shared backbone and compare direct generation against \textsc{IBA-Agent} on the same task and with the same preference updates. 
In the writing case, \textsc{IBA-Agent} correctly applies the user's latest preference to open with a personal story, whereas the baseline follows the outdated preference and opens with a viewpoint.
This suggests that \textsc{IBA-Agent} can better isolate dynamic preference cues and synthesize them into actions.
However, in the diet-planning case, both methods fail to update the plan from three meals to four meals per day. 
This indicates that the task remains difficult when the updated preference requires a global structural revision of the output, rather than a localized stylistic change.

\section{Conclusion}
In this paper, we identified a critical knowledge-to-action gap in existing personalized agent evaluations, which rely heavily on static user profiles and explicit question answering.
We argued that true personalization requires agents to synthesize implicit signals from longitudinal histories rather than merely retrieving explicit tags.
To address this, we introduced \textsc{IBA-Bench}, a comprehensive benchmark that evaluates implicit behavioral alignment under dynamic and noisy conditions.
We further proposed IBA-Agent, a framework that bridges information retrieval and task execution through deep retrieval and planning-guided generation.
Extensive experiments demonstrate that while current state-of-the-art models struggle to translate historical context into behavioral constraints, our approach significantly improves performance in these complex scenarios.
Closing this gap is a prerequisite for personalized agents in the wild.

\section*{Limitations}
We propose \textsc{IBA-Bench} as a benchmark for evaluating implicit behavioral alignment in personalized agents; however, it still has two main limitations. First, \textsc{IBA-Bench} is constructed using LLM-generated synthetic data. Although the synthesis pipeline is carefully designed to inject implicit preferences, temporal dynamics, and noise, hallucinated or inconsistent behaviors may still occur. 
Second, our evaluation focuses on offline, history-conditioned task execution.
As a result, the benchmark cannot capture how agents adapt preferences through online interaction or corrective signals. Extending \textsc{IBA-Bench} to interactive and tool-augmented settings is left for future work.

\section*{Ethical Considerations}
Considering potential risks related to privacy and intellectual property, we deliberately avoid using real user logs or information collected from human subjects. All personas, interaction histories, and tasks are fictional, thereby preventing direct exposure of personally identifiable information. At the same time, as LLM-generated content may still reflect biased or undesirable patterns, we address these issues through carefully designed prompts and targeted human review.
The benchmark data and prompts will be released under CC BY-NC 4.0, and the evaluation scripts will be released under the MIT License. These released materials are intended for academic research; any redistribution or derivative use should follow the stated licenses and remain within this research and evaluation scope.

\FloatBarrier
\bibliography{custom}

@article{openai2024gpt4technicalreport,
  title={Gpt-4 technical report},
  author={Achiam, Josh and Adler, Steven and Agarwal, Sandhini and Ahmad, Lama and Akkaya, Ilge and Aleman, Florencia Leoni and Almeida, Diogo and Altenschmidt, Janko and Altman, Sam and Anadkat, Shyamal and others},
  journal={arXiv preprint arXiv:2303.08774},
  year={2023}
}

@article{comanici2025gemini,
  author       = {Gemini Team},
  title        = {Gemini 2.5: Pushing the Frontier with Advanced Reasoning, Multimodality,
                  Long Context, and Next Generation Agentic Capabilities},
  journal      = {CoRR},
  volume       = {abs/2507.06261},
  year         = {2025},
  url          = {https://doi.org/10.48550/arXiv.2507.06261},
  doi          = {10.48550/ARXIV.2507.06261},
  eprinttype   = {arXiv},
  eprint       = {2507.06261},
  bibsource    = {dblp computer science bibliography, https://dblp.org}
}

@article{guo2025DeepSeek,
  author       = {Daya Guo and
                  Dejian Yang and
                  Haowei Zhang and
                  Junxiao Song and
                  Peiyi Wang and
                  Qihao Zhu and
                  Runxin Xu and
                  Ruoyu Zhang and
                  Shirong Ma and
                  Xiao Bi and
                  Xiaokang Zhang and
                  Xingkai Yu and
                  Yu Wu and
                  Z. F. Wu and
                  Zhibin Gou and
                  Zhihong Shao and
                  Zhuoshu Li and
                  Ziyi Gao and
                  Aixin Liu and
                  Bing Xue and
                  Bingxuan Wang and
                  Bochao Wu and
                  Bei Feng and
                  Chengda Lu and
                  Chenggang Zhao and
                  Chengqi Deng and
                  Chong Ruan and
                  Damai Dai and
                  Deli Chen and
                  Dongjie Ji and
                  Erhang Li and
                  Fangyun Lin and
                  Fucong Dai and
                  Fuli Luo and
                  Guangbo Hao and
                  Guanting Chen and
                  Guowei Li and
                  Hao Zhang and
                  Hanwei Xu and
                  Honghui Ding and
                  Huazuo Gao and
                  Hui Qu and
                  Hui Li and
                  Jianzhong Guo and
                  Jiashi Li and
                  Jingchang Chen and
                  Jingyang Yuan and
                  Jinhao Tu and
                  Junjie Qiu and
                  Junlong Li and
                  J. L. Cai and
                  Jiaqi Ni and
                  Jian Liang and
                  Jin Chen and
                  Kai Dong and
                  Kai Hu and
                  Kaichao You and
                  Kaige Gao and
                  Kang Guan and
                  Kexin Huang and
                  Kuai Yu and
                  Lean Wang and
                  Lecong Zhang and
                  Liang Zhao and
                  Litong Wang and
                  Liyue Zhang and
                  Lei Xu and
                  Leyi Xia and
                  Mingchuan Zhang and
                  Minghua Zhang and
                  Minghui Tang and
                  Mingxu Zhou and
                  Meng Li and
                  Miaojun Wang and
                  Mingming Li and
                  Ning Tian and
                  Panpan Huang and
                  Peng Zhang and
                  Qiancheng Wang and
                  Qinyu Chen and
                  Qiushi Du and
                  Ruiqi Ge and
                  Ruisong Zhang and
                  Ruizhe Pan and
                  Runji Wang and
                  R. J. Chen and
                  R. L. Jin and
                  Ruyi Chen and
                  Shanghao Lu and
                  Shangyan Zhou and
                  Shanhuang Chen and
                  Shengfeng Ye and
                  Shiyu Wang and
                  Shuiping Yu and
                  Shunfeng Zhou and
                  Shuting Pan and
                  S. S. Li and
                  Shuang Zhou and
                  Shaoqing Wu and
                  Tao Yun and
                  Tian Pei and
                  Tianyu Sun and
                  Tao Wang and
                  Wangding Zeng and
                  Wen Liu and
                  Wenfeng Liang and
                  Wenjun Gao and
                  Wenqin Yu and
                  Wentao Zhang and
                  W. L. Xiao and
                  Wei An and
                  Xiaodong Liu and
                  Xiaohan Wang and
                  Xiaokang Chen and
                  Xiaotao Nie and
                  Xin Cheng and
                  Xin Liu and
                  Xin Xie and
                  Xingchao Liu and
                  Xinyu Yang and
                  Xinyuan Li and
                  Xuecheng Su and
                  Xuheng Lin and
                  X. Q. Li and
                  Xiangyue Jin and
                  Xiaojin Shen and
                  Xiaosha Chen and
                  Xiaowen Sun and
                  Xiaoxiang Wang and
                  Xinnan Song and
                  Xinyi Zhou and
                  Xianzu Wang and
                  Xinxia Shan and
                  Y. K. Li and
                  Y. Q. Wang and
                  Y. X. Wei and
                  Yang Zhang and
                  Yanhong Xu and
                  Yao Li and
                  Yao Zhao and
                  Yaofeng Sun and
                  Yaohui Wang and
                  Yi Yu and
                  Yichao Zhang and
                  Yifan Shi and
                  Yiliang Xiong and
                  Ying He and
                  Yishi Piao and
                  Yisong Wang and
                  Yixuan Tan and
                  Yiyang Ma and
                  Yiyuan Liu and
                  Yongqiang Guo and
                  Yuan Ou and
                  Yuduan Wang and
                  Yue Gong and
                  Yuheng Zou and
                  Yujia He and
                  Yunfan Xiong and
                  Yuxiang Luo and
                  Yuxiang You and
                  Yuxuan Liu and
                  Yuyang Zhou and
                  Y. X. Zhu and
                  Yanping Huang and
                  Yaohui Li and
                  Yi Zheng and
                  Yuchen Zhu and
                  Yunxian Ma and
                  Ying Tang and
                  Yukun Zha and
                  Yuting Yan and
                  Z. Z. Ren and
                  Zehui Ren and
                  Zhangli Sha and
                  Zhe Fu and
                  Zhean Xu and
                  Zhenda Xie and
                  Zhengyan Zhang and
                  Zhewen Hao and
                  Zhicheng Ma and
                  Zhigang Yan and
                  Zhiyu Wu and
                  Zihui Gu and
                  Zijia Zhu and
                  Zijun Liu and
                  Zilin Li and
                  Ziwei Xie and
                  Ziyang Song and
                  Zizheng Pan and
                  Zhen Huang and
                  Zhipeng Xu and
                  Zhongyu Zhang and
                  Zhen Zhang},
  title        = {DeepSeek-R1 incentivizes reasoning in LLMs through reinforcement learning},
  journal      = {Nat.},
  volume       = {645},
  number       = {8081},
  pages        = {633--638},
  year         = {2025},
  url          = {https://doi.org/10.1038/s41586-025-09422-z},
  doi          = {10.1038/S41586-025-09422-Z},
  bibsource    = {dblp computer science bibliography, https://dblp.org}
}

@inproceedings{yang18hotpotqa,
  author       = {Zhilin Yang and
                  Peng Qi and
                  Saizheng Zhang and
                  Yoshua Bengio and
                  William W. Cohen and
                  Ruslan Salakhutdinov and
                  Christopher D. Manning},
  title        = {HotpotQA: {A} Dataset for Diverse, Explainable Multi-hop Question
                  Answering},
  booktitle    = {Proceedings of EMNLP},
  pages        = {2369--2380},
  year         = {2018},
}

@inproceedings{yang24sweagent,
  author       = {John Yang and
                  Carlos E. Jimenez and
                  Alexander Wettig and
                  Kilian Lieret and
                  Shunyu Yao and
                  Karthik Narasimhan and
                  Ofir Press},
  title        = {SWE-agent: Agent-Computer Interfaces Enable Automated Software Engineering},
  booktitle    = {NeurIPS},
  year         = {2024},
}

@inproceedings{zhao25personalens,
  author       = {Zheng Zhao and
                  Clara Vania and
                  Subhradeep Kayal and
                  Naila Khan and
                  Shay B. Cohen and
                  Emine Yilmaz},
  title        = {PersonaLens: {A} Benchmark for Personalization Evaluation in Conversational
                  {AI} Assistants},
  booktitle    = {Findings of ACL},
  pages        = {18023--18055},
  year         = {2025},
}

@inproceedings{samuel24personagym,
  author       = {Vinay Samuel and
                  Henry Peng Zou and
                  Yue Zhou and
                  Shreyas Chaudhari and
                  Ashwin Kalyan and
                  Tanmay Rajpurohit and
                  Ameet Deshpande and
                  Karthik R. Narasimhan and
                  Vishvak Murahari},
  title        = {PersonaGym: Evaluating Persona Agents and LLMs},
  booktitle    = {Findings of EMNLP},
  pages        = {6999--7022},
  year         = {2025},
}

@article{zhang25personaagent,
  title={Personaagent: When large language model agents meet personalization at test time},
  author={Zhang, Weizhi and Zhang, Xinyang and Zhang, Chenwei and Yang, Liangwei and Shang, Jingbo and Wei, Zhepei and Zou, Henry Peng and Huang, Zijie and Wang, Zhengyang and Gao, Yifan and others},
  journal={arXiv preprint arXiv:2506.06254},
  year={2025}
}

@inproceedings{joko24dpl,
  author       = {Hideaki Joko and
                  Shubham Chatterjee and
                  Andrew Ramsay and
                  Arjen P. de Vries and
                  Jeff Dalton and
                  Faegheh Hasibi},
  title        = {Doing Personal {LAPS:} LLM-Augmented Dialogue Construction for Personalized
                  Multi-Session Conversational Search},
  booktitle    = {Proceedings of ACM SIGIR},
  pages        = {796--806},
  year         = {2024},
}

@inproceedings{mok25etp,
  author       = {Jisoo Mok and
                  Ik{-}hwan Kim and
                  Sangkwon Park and
                  Sungroh Yoon},
  title        = {Exploring the Potential of LLMs as Personalized Assistants: Dataset,
                  Evaluation, and Analysis},
  booktitle    = {Proceedings of ACL},
  pages        = {10212--10239},
  year         = {2025},
}

@inproceedings{tan25personabench,
  author       = {Juntao Tan and
                  Liangwei Yang and
                  Zuxin Liu and
                  Zhiwei Liu and
                  Rithesh R. N. and
                  Tulika Manoj Awalgaonkar and
                  Jianguo Zhang and
                  Weiran Yao and
                  Ming Zhu and
                  Shirley Kokane and
                  Silvio Savarese and
                  Huan Wang and
                  Caiming Xiong and
                  Shelby Heinecke},
  title        = {PersonaBench: Evaluating {AI} Models on Understanding Personal Information
                  through Accessing (Synthetic) Private User Data},
  booktitle    = {Findings of ACL},
  pages        = {878--893},
  year         = {2025},
}

@misc{qwen2025qwen25technicalreport,
      title={Qwen2.5 Technical Report}, 
      author={Qwen and An Yang and Baosong Yang and Beichen Zhang and Binyuan Hui and Bo Zheng and Bowen Yu and Chengyuan Li and Dayiheng Liu and Fei Huang and Haoran Wei and Huan Lin and Jian Yang and Jianhong Tu and Jianwei Zhang and Jianxin Yang and Jiaxi Yang and Jingren Zhou and Junyang Lin and Kai Dang and Keming Lu and Keqin Bao and Kexin Yang and Le Yu and Mei Li and Mingfeng Xue and Pei Zhang and Qin Zhu and Rui Men and Runji Lin and Tianhao Li and Tianyi Tang and Tingyu Xia and Xingzhang Ren and Xuancheng Ren and Yang Fan and Yang Su and Yichang Zhang and Yu Wan and Yuqiong Liu and Zeyu Cui and Zhenru Zhang and Zihan Qiu},
      year={2025},
      eprint={2412.15115},
      archivePrefix={arXiv},
      primaryClass={cs.CL},
      url={https://arxiv.org/abs/2412.15115}, 
}

@article{glm2024chatglmfamilylargelanguage,
  title={Chatglm: A family of large language models from glm-130b to glm-4 all tools},
  author={Glm, Team and Zeng, Aohan and Xu, Bin and Wang, Bowen and Zhang, Chenhui and Yin, Da and Zhang, Dan and Rojas, Diego and Feng, Guanyu and Zhao, Hanlin and others},
  journal={arXiv preprint arXiv:2406.12793},
  year={2024}
}

@article{deepseekai2025deepseekv3technicalreport,
  title={Deepseek-v3 technical report},
  author={Liu, Aixin and Feng, Bei and Xue, Bing and Wang, Bingxuan and Wu, Bochao and Lu, Chengda and Zhao, Chenggang and Deng, Chengqi and Zhang, Chenyu and Ruan, Chong and others},
  journal={arXiv preprint arXiv:2412.19437},
  year={2024}
}

@article{deepseekai2025deepseekv32pushingfrontieropen,
  title={Deepseek-v3. 2: Pushing the frontier of open large language models},
  author={Liu, Aixin and Mei, Aoxue and Lin, Bangcai and Xue, Bing and Wang, Bingxuan and Xu, Bingzheng and Wu, Bochao and Zhang, Bowei and Lin, Chaofan and Dong, Chen and others},
  journal={arXiv preprint arXiv:2512.02556},
  year={2025}
}

@inproceedings{hao25evaluating,
  author       = {Yupu Hao and
                  Pengfei Cao and
                  Zhuoran Jin and
                  Huanxuan Liao and
                  Yubo Chen and
                  Kang Liu and
                  Jun Zhao},
  title        = {Evaluating Personalized Tool-Augmented LLMs from the Perspectives
                  of Personalization and Proactivity},
  booktitle    = {Proceedings of ACL},
  pages        = {21897--21935},
  year         = {2025},
}

@article{westhausser25longterm,
  title={Enabling personalized long-term interactions in llm-based agents through persistent memory and user profiles},
  author={Westh{\"a}u{\ss}er, Rebecca and Minker, Wolfgang and Zepf, Sebastian},
  journal={arXiv preprint arXiv:2510.07925},
  year={2025}
}

@inproceedings{liu2025agentbenchevaluatingllmsagents,
  author       = {Xiao Liu and
                  Hao Yu and
                  Hanchen Zhang and
                  Yifan Xu and
                  Xuanyu Lei and
                  Hanyu Lai and
                  Yu Gu and
                  Hangliang Ding and
                  Kaiwen Men and
                  Kejuan Yang and
                  Shudan Zhang and
                  Xiang Deng and
                  Aohan Zeng and
                  Zhengxiao Du and
                  Chenhui Zhang and
                  Sheng Shen and
                  Tianjun Zhang and
                  Yu Su and
                  Huan Sun and
                  Minlie Huang and
                  Yuxiao Dong and
                  Jie Tang},
  title        = {AgentBench: Evaluating LLMs as Agents},
  booktitle    = {The Twelfth ICLR},
  year         = {2024},
}

@inproceedings{10.1145/3586183.3606763,
  author       = {Joon Sung Park and
                  Joseph C. O'Brien and
                  Carrie Jun Cai and
                  Meredith Ringel Morris and
                  Percy Liang and
                  Michael S. Bernstein},
  title        = {Generative Agents: Interactive Simulacra of Human Behavior},
  booktitle    = {Proceedings of UIST},
  pages        = {2:1--2:22},
  year         = {2023},
}

@inproceedings{wei2022chain,
  author       = {Jason Wei and
                  Xuezhi Wang and
                  Dale Schuurmans and
                  Maarten Bosma and
                  Brian Ichter and
                  Fei Xia and
                  Ed H. Chi and
                  Quoc V. Le and
                  Denny Zhou},
  title        = {Chain-of-Thought Prompting Elicits Reasoning in Large Language Models},
  booktitle    = {NeurIPS},
  year         = {2022},
}

@inproceedings{valmeekam2023planning,
  author       = {Karthik Valmeekam and
                  Matthew Marquez and
                  Sarath Sreedharan and
                  Subbarao Kambhampati},
  title        = {On the Planning Abilities of Large Language Models - {A} Critical
                  Investigation},
  booktitle    = {NeurIPS},
  year         = {2023},
}

@inproceedings{schick2023toolformer,
  author       = {Timo Schick and
                  Jane Dwivedi{-}Yu and
                  Roberto Dess{\`{\i}} and
                  Roberta Raileanu and
                  Maria Lomeli and
                  Eric Hambro and
                  Luke Zettlemoyer and
                  Nicola Cancedda and
                  Thomas Scialom},
  title        = {Toolformer: Language Models Can Teach Themselves to Use Tools},
  booktitle    = {NeurIPS},
  year         = {2023},
}

@article{wang2024aipersonalifelongpersonalization,
  title={Ai persona: Towards life-long personalization of llms},
  author={Wang, Tiannan and Tao, Meiling and Fang, Ruoyu and Wang, Huilin and Wang, Shuai and Jiang, Yuchen Eleanor and Zhou, Wangchunshu},
  journal={arXiv preprint arXiv:2412.13103},
  year={2024}
}

@article{ram2023context,
  title={In-context retrieval-augmented language models},
  author={Ram, Ori and Levine, Yoav and Dalmedigos, Itay and Muhlgay, Dor and Shashua, Amnon and Leyton-Brown, Kevin and Shoham, Yoav},
  journal={Transactions of the Association for Computational Linguistics},
  volume={11},
  pages={1316--1331},
  year={2023},
  publisher={MIT Press One Broadway, 12th Floor, Cambridge, Massachusetts 02142, USA~…}
}

@article{stiennon2022learningsummarizehumanfeedback,
  title={Learning to summarize with human feedback},
  author={Stiennon, Nisan and Ouyang, Long and Wu, Jeffrey and Ziegler, Daniel and Lowe, Ryan and Voss, Chelsea and Radford, Alec and Amodei, Dario and Christiano, Paul F},
  journal={Advances in neural information processing systems},
  volume={33},
  pages={3008--3021},
  year={2020}
}

@misc{QwQ32BBlog2025,
  title        = {QwQ-32B: Embracing the Power of Reinforcement Learning},
  author       = {{Qwen Team}},
  year         = {2025},
  month        = {Mar 6},
  howpublished = {\url{https://qwenlm.github.io/blog/qwq-32b/}},
  note         = {Official model introduction blog. Accessed: 2026-01-04},
  url          = {https://qwenlm.github.io/blog/qwq-32b/}
}

@inproceedings{10.1145/3711896.3736570,
  author       = {Mahmoud Mohammadi and
                  Yipeng Li and
                  Jane Lo and
                  Wendy Yip},
  title        = {Evaluation and Benchmarking of {LLM} Agents: {A} Survey},
  booktitle    = {Proceedings of ACM SIGKDD},
  pages        = {6129--6139},
  year         = {2025},
}

@misc{OpenAI_GPT4o2024,
  title        = {Hello GPT-4o},
  author       = {{OpenAI}},
  year         = {2024},
  month        = {May 13},
  howpublished = {\url{https://openai.com/index/hello-gpt-4o/}},
  note         = {Official announcement of GPT-4o model; accessed: 2026-01-04},
  url          = {https://openai.com/index/hello-gpt-4o/}
}

@misc{OpenAI_GPT5_1_2025,
  title        = {GPT‑5.1: A smarter, more conversational ChatGPT},
  author       = {{OpenAI}},
  year         = {2025},
  month        = {Nov 12},
  howpublished = {\url{https://openai.com/index/gpt-5-1/}},
  note         = {Official announcement of GPT-5.1 model series; accessed: 2026-01-04},
  url          = {https://openai.com/index/gpt-5-1/}
}

@article{hu2023languagemodelsagentmodels,
  title={Language models, agent models, and world models: The law for machine reasoning and planning},
  author={Hu, Zhiting and Shu, Tianmin},
  journal={arXiv preprint arXiv:2312.05230},
  year={2023}
}

@article{zhu2025evolutionaryperspectivesevaluationllmbased,
  title={Evolutionary Perspectives on the Evaluation of LLM-Based AI Agents: A Comprehensive Survey},
  author={Zhu, Jiachen and Zhu, Menghui and Rui, Renting and Shan, Rong and Zheng, Congmin and Chen, Bo and Xi, Yunjia and Lin, Jianghao and Liu, Weiwen and Tang, Ruiming and others},
  journal={arXiv preprint arXiv:2506.11102},
  year={2025}
}

@article{qwen3technicalreport,
  title={Qwen3 technical report},
  author={Yang, An and Li, Anfeng and Yang, Baosong and Zhang, Beichen and Hui, Binyuan and Zheng, Bo and Yu, Bowen and Gao, Chang and Huang, Chengen and Lv, Chenxu and others},
  journal={arXiv preprint arXiv:2505.09388},
  year={2025}
}

@article{gwet2001handbook,
  title={Handbook of inter-rater reliability: How to estimate the level of agreement between two or multiple raters},
  author={Gwet, Kilem},
  journal={Gaithersburg, MD: STATAXIS Publishing Company},
  year={2001}
}

@misc{ren25nanobot,
  author       = {Ren, Xubin and {HKUDS}},
  title        = {nanobot: The Ultra-Lightweight Personal {AI} Agent},
  year         = {2026},
  howpublished = {\url{https://github.com/HKUDS/nanobot}},
  note         = {GitHub repository, accessed May 2026}
}

@misc{nous25hermes,
  author       = {{Nous Research}},
  title        = {Hermes Agent: The Agent That Grows With You},
  year         = {2026},
  howpublished = {\url{https://github.com/NousResearch/hermes-agent}},
  note         = {GitHub repository, accessed May 2026}
}

@article{afz24corr,
  author       = {Saleh Afzoon and
                  Usman Naseem and
                  Amin Beheshti and
                  Zahra Jamali},
  title        = {PersoBench: Benchmarking Personalized Response Generation in Large
                  Language Models},
  journal      = {CoRR},
  volume       = {abs/2410.03198},
  year         = {2024},
}

@misc{li2026for,
      title={FORTIS: Benchmarking Over-Privilege in Agent Skills}, 
      author={Shawn Li and Chenxiao Yu and Han Wang and Wei Yang and Ryan Rossi and Franck Dernoncourt and Xiyang Hu and Philip Yu and Chaowei Xiao and Huan Zhang and Yue Zhao},
      year={2026},
      eprint={2605.09163},
      archivePrefix={arXiv},
      primaryClass={cs.AI},
      url={https://arxiv.org/abs/2605.09163}, 
}

\appendix
\section*{Appendix}

\begin{table*}[!ht]
\centering
\footnotesize
\setlength{\tabcolsep}{5.5pt}
\renewcommand{\arraystretch}{1.08}
\resizebox{1.0\linewidth}{!}{
\begin{tabular}{l *{10}{c}}
\toprule
\textbf{Model} & \textbf{Writing} & \textbf{Work} & \textbf{Daily} & \textbf{Plan.} & \textbf{Health} & \textbf{Trans.} & \textbf{Med.} & \textbf{Leis.} & \textbf{Info.} & \textbf{Overall} \\
\midrule

\specialrule{0.08em}{0.15em}{0.15em}
\multicolumn{11}{c}{\textbf{\textit{$\bullet$ Baseline Models}}} \\
\specialrule{0.08em}{0.15em}{0.15em}
Qwen2.5-7B-Instruct~\cite{qwen2025qwen25technicalreport} & 49.3 & 49.5 & 77.0 & 56.7 & 68.1 & 68.3 & 72.9 & 40.2 & 33.7 & 57.8 \\
DeepSeek-V3~\cite{deepseekai2025deepseekv3technicalreport} & 64.1 & 57.2 & 77.7 & 67.0 & 74.2 & 71.7 & 79.3 & 50.2 & 51.2 & 66.5 \\
ChatGLM-4-9b-chat~\cite{glm2024chatglmfamilylargelanguage} & 45.6 & 45.2 & 70.9 & 54.3 & 66.2 & 60.6 & 69.0 & 38.0 & 33.0 & 54.4 \\
QwQ-32B~\cite{QwQ32BBlog2025} & 67.2 & 62.2 & 81.3 & 68.9 & 77.9 & 79.0 & 80.3 & 53.4 & 56.0 & 70.2 \\
GPT-4o-mini & 54.7 & 54.1 & 78.2 & 61.3 & 69.0 & 71.5 & 71.7 & 45.2 & 39.2 & 61.6 \\
GPT-4o~\cite{OpenAI_GPT4o2024} & 58.8 & 53.0 & 76.7 & 61.3 & 70.5 & 70.6 & 73.1 & 49.1 & 39.8 & 62.0 \\
GPT-5-mini & 73.2 & 66.0 & 76.4 & 75.7 & 81.9 & 74.1 & 81.9 & 52.4 & 68.0 & 73.1 \\
GPT-5.1~\cite{OpenAI_GPT5_1_2025} & 75.8 & 63.2 & 80.2 & 70.2 & 75.5 & 74.0 & 78.8 & 58.0 & 59.1 & 71.1 \\

\midrule
\specialrule{0.08em}{0.15em}{0.15em}
\multicolumn{11}{c}{\textbf{\textit{$\bullet$ DeepSeek-V3.2 Variants}}} \\
\specialrule{0.08em}{0.15em}{0.15em}
DeepSeek-V3.2~\cite{deepseekai2025deepseekv32pushingfrontieropen} & 69.4 & 61.3 & 80.3 & 67.6 & 73.7 & 75.4 & 79.6 & 54.1 & 55.8 & 69.1 \\
IBA-Agent (DeepSeek-V3.2) & \uline{86.3} & \uline{78.3} & \uline{85.3} & 82.1 & 86.5 & 80.7 & 87.3 & \best{60.3} & \best{79.9} & \uline{81.9} \\
\rowcolor{lightpurple}
IBA-Agent (DeepSeek-V3.2, FC) & \best{88.3} & \best{79.6} & \best{89.2} & \uline{82.9} & \best{90.4} & \best{84.0} & \best{92.2} & \uline{59.8} & \uline{78.7} & \best{84.0} \\

\midrule
\specialrule{0.08em}{0.15em}{0.15em}
\multicolumn{11}{c}{\textbf{\textit{$\bullet$ Qwen3-4B-Instruct Variants}}} \\
\specialrule{0.08em}{0.15em}{0.15em}
Qwen3-4B-Instruct~\cite{qwen3technicalreport} & 63.6 & 59.5 & 80.1 & 66.1 & 74.7 & 76.0 & 81.5 & 48.9 & 53.9 & 67.8 \\
IBA-Agent (Qwen3-4B-Instruct) & 82.4 & 75.1 & 84.4 & 81.7 & 87.5 & \uline{82.5} & 88.8 & 49.9 & 73.6 & 80.0 \\
\rowcolor{lightpurple}
IBA-Agent (Qwen3-4B-Instruct, FC) & 81.1 & 76.0 & 83.1 & \best{83.5} & \uline{88.5} & \uline{82.5} & \uline{90.0} & 50.0 & 72.0 & 80.2 \\

\midrule
\specialrule{0.08em}{0.15em}{0.15em}
\multicolumn{11}{c}{\textbf{\textit{$\bullet$ General-Purpose AI Agents}}} \\
\specialrule{0.08em}{0.15em}{0.15em}
nanobot (DeepSeek-V3.2) & 69.7 & 52.7 & 60.4 & 67.4 & 61.1 & 59.4 & 60.4 & 44.8 & 67.1 & 61.2 \\
Claude Code (DeepSeek-V3.2) & 75.1 & 59.5 & 60.3 & 69.5 & 62.3 & 59.1 & 57.1 & 38.7 & 57.3 & 61.8 \\
Hermes Agent (DeepSeek-V3.2) & 79.1 & 60.2 & 61.3 & 71.7 & 62.8 & 59.3 & 60.3 & 41.7 & 69.6 & 64.5  \\

\bottomrule
\end{tabular}
}
\caption{Performance comparison using the \textbf{bge-m3} retriever and \textbf{Qwen3-30B-A3B-Instruct} as the judge model. The columns correspond to nine application domains: \textbf{Writing} (writing and design), \textbf{Work}, \textbf{Daily} (daily consumption), \textbf{Plan.} (planning), \textbf{Health} (exercise and health), \textbf{Trans.} (transportation), \textbf{Med.} (medical services), \textbf{Leis.} (leisure activities), and \textbf{Info.} (information management). The best result in each column is highlighted as \colorbox{bestgreen}{\textbf{green bold text}}, the second-best result is \uline{underlined}, \colorbox{lightpurple}{Function Call variants} are shaded in light purple. FC stands for Function Call.}
\label{tab:res_30b}
\end{table*}

\begin{table*}[!ht]
\centering
\footnotesize
\setlength{\tabcolsep}{5.5pt}
\renewcommand{\arraystretch}{1.08}
\resizebox{1.0\linewidth}{!}{
\begin{tabular}{l *{10}{c}}
\toprule
\textbf{Model} & \textbf{Writing} & \textbf{Work} & \textbf{Daily} & \textbf{Plan.} & \textbf{Health} & \textbf{Trans.} & \textbf{Med.} & \textbf{Leis.} & \textbf{Info.} & \textbf{Overall} \\
\midrule

\specialrule{0.08em}{0.15em}{0.15em}
\multicolumn{11}{c}{\textbf{\textit{$\bullet$ Baseline Models}}} \\
\specialrule{0.08em}{0.15em}{0.15em}
Qwen2.5-7B-Instruct~\cite{qwen2025qwen25technicalreport} & 42.7 & 42.0 & 75.0 & 45.8 & 57.9 & 60.7 & 73.9 & 32.8 & 32.4 & 51.0 \\
DeepSeek-V3~\cite{deepseekai2025deepseekv3technicalreport} & 48.2 & 43.2 & 74.7 & 50.3 & 61.2 & 62.2 & 75.6 & 34.6 & 40.1 & 55.0 \\
ChatGLM-4-9b-chat~\cite{glm2024chatglmfamilylargelanguage} & 42.1 & 39.0 & 67.2 & 43.8 & 56.5 & 54.7 & 67.1 & 29.8 & 31.8 & 48.5 \\
QwQ-32B~\cite{QwQ32BBlog2025} & 54.5 & 50.0 & 77.2 & 54.1 & 67.1 & 69.3 & 76.3 & 39.3 & 43.1 & 59.6 \\
GPT-4o-mini & 47.1 & 44.2 & 74.2 & 47.1 & 59.7 & 64.1 & 69.4 & 30.7 & 34.2 & 53.2 \\
GPT-4o~\cite{OpenAI_GPT4o2024} & 55.1 & 42.0 & 72.0 & 45.2 & 60.8 & 62.3 & 69.9 & 35.9 & 33.9 & 53.3 \\
GPT-5-mini & 57.3 & 52.4 & 73.4 & 58.6 & 71.1 & 66.1 & 76.6 & 42.5 & 48.1 & 61.3 \\
GPT-5.1~\cite{OpenAI_GPT5_1_2025} & 58.4 & 49.0 & 76.2 & 54.2 & 66.5 & 65.7 & 75.8 & \uline{49.0} & 43.4 & 59.8 \\

\midrule
\specialrule{0.08em}{0.15em}{0.15em}
\multicolumn{11}{c}{\textbf{\textit{$\bullet$ DeepSeek-V3.2 Variants}}} \\
\specialrule{0.08em}{0.15em}{0.15em}
DeepSeek-V3.2~\cite{deepseekai2025deepseekv32pushingfrontieropen} & 54.6 & 47.2 & 76.9 & 52.5 & 64.4 & 64.7 & 77.0 & 39.3 & 40.5 & 58.0 \\
IBA-Agent (DeepSeek-V3.2) & \uline{72.2} & \uline{64.8} & \uline{83.5} & \uline{69.3} & \uline{82.5} & 73.4 & 87.1 & 42.5 & 57.0 & \uline{71.8} \\
\rowcolor{lightpurple}
IBA-Agent (DeepSeek-V3.2, FC) & \best{78.4} & \best{69.3} & \best{87.8} & \best{75.3} & \best{88.1} & 76.3 & \best{92.4} & 43.9 & \best{62.6} & \best{76.7} \\

\midrule
\specialrule{0.08em}{0.15em}{0.15em}
\multicolumn{11}{c}{\textbf{\textit{$\bullet$ Qwen3-4B-Instruct Variants}}} \\
\specialrule{0.08em}{0.15em}{0.15em}
Qwen3-4B-Instruct~\cite{qwen3technicalreport} & 52.3 & 49.4 & 71.0 & 51.9 & 65.5 & \best{88.0} & 82.9 & \best{59.5} & 43.7 & 60.6 \\
IBA-Agent (Qwen3-4B-Instruct) & 71.9 & 65.2 & 82.4 & 71.0 & 82.9 & \uline{80.2} & 89.6 & 39.6 & \uline{58.0} & 72.6 \\
\rowcolor{lightpurple}
IBA-Agent (Qwen3-4B-Instruct, FC) & 72.6 & 65.1 & 80.6 & 72.4 & 84.1 & 79.9 & \uline{91.1} & 36.3 & 60.6 & 73.0 \\

\midrule
\specialrule{0.08em}{0.15em}{0.15em}
\multicolumn{11}{c}{\textbf{\textit{$\bullet$ General-Purpose AI Agents}}} \\
\specialrule{0.08em}{0.15em}{0.15em}
nanobot (DeepSeek-V3.2) & 58.4 & 44.8 & 54.3 & 51.9 & 45.6 & 45.5 & 56.9 & 28.4 & 45.8 & 49.3 \\
Claude Code (DeepSeek-V3.2) & 81.5 & 65.4 & 56.9 & 67.6 & 58.4 & 50.5 & 56.4 & 26.8 & 56.5 & 60.9 \\
Hermes Agent (DeepSeek-V3.2) & 82.1 & 63.4 & 35.2 & 67.8 & 48.6 & 54.6 & 55.5 & 28.2 & 67.1 & 57.5 \\
\bottomrule
\end{tabular}
}
\caption{Performance comparison using the \textbf{bge-m3} retriever and \textbf{DeepSeek-V3.2} as the judge model. The columns correspond to nine application domains: \textbf{Writing} (writing and design), \textbf{Work}, \textbf{Daily} (daily consumption), \textbf{Plan.} (planning), \textbf{Health} (exercise and health), \textbf{Trans.} (transportation), \textbf{Med.} (medical services), \textbf{Leis.} (leisure activities), and \textbf{Info.} (information management). The best result in each column is highlighted as \colorbox{bestgreen}{\textbf{green bold text}}, the second-best result is \uline{underlined}, \colorbox{lightpurple}{Function Call variants} are shaded in light purple. FC stands for Function Call.}
\label{tab:res_ds}
\end{table*}

\begin{table*}[!ht]
\centering
\footnotesize
\setlength{\tabcolsep}{5.5pt}
\renewcommand{\arraystretch}{1.08}
\resizebox{1.0\linewidth}{!}{
\begin{tabular}{l *{10}{c}}
\toprule
\textbf{Model} & \textbf{Writing} & \textbf{Work} & \textbf{Daily} & \textbf{Plan.} & \textbf{Health} & \textbf{Trans.} & \textbf{Med.} & \textbf{Leis.} & \textbf{Info.} & \textbf{Overall} \\
\midrule

\specialrule{0.08em}{0.15em}{0.15em}
\multicolumn{11}{c}{\textbf{\textit{$\bullet$ Baseline Models}}} \\
\specialrule{0.08em}{0.15em}{0.15em}
Qwen2.5-7B-Instruct~\cite{qwen2025qwen25technicalreport} & 66.0 & 61.5 & 79.8 & 64.0 & 75.2 & 66.5 & 78.2 & 47.3 & 45.3 & 66.0 \\
DeepSeek-V3~\cite{deepseekai2025deepseekv3technicalreport} & 83.1 & 69.3 & 78.8 & 70.8 & 78.8 & 68.1 & 81.4 & 49.2 & 64.4 & 73.1 \\
ChatGLM-4-9b-chat~\cite{glm2024chatglmfamilylargelanguage} & 66.6 & 57.6 & 74.0 & 62.3 & 73.4 & 60.0 & 71.7 & 42.3 & 46.5 & 63.1 \\
QwQ-32B~\cite{QwQ32BBlog2025} & 84.4 & 72.6 & 83.4 & 70.4 & 80.6 & 74.4 & 82.1 & 49.3 & 67.0 & 75.3 \\
GPT-4o-mini & 71.3 & 66.4 & 79.9 & 65.2 & 75.6 & 69.2 & 74.3 & 43.7 & 53.7 & 68.4 \\
GPT-4o~\cite{OpenAI_GPT4o2024} & 76.5 & 67.4 & 78.0 & 67.2 & 77.3 & 67.0 & 75.9 & 47.4 & 56.7 & 69.7 \\
GPT-5-mini & 81.02 & 73.34 & 79.20 & 75.16 & 82.12 & 72.32 & 82.26 & 54.78 & 73.58 & 76.10 \\
GPT-5.1~\cite{OpenAI_GPT5_1_2025} & 84.6 & 75.0 & 83.2 & 72.5 & 79.7 & 72.2 & 79.5 & \best{63.3} & 71.9 & 76.7 \\

\midrule
\specialrule{0.08em}{0.15em}{0.15em}
\multicolumn{11}{c}{\textbf{\textit{$\bullet$ DeepSeek-V3.2 Variants}}} \\
\specialrule{0.08em}{0.15em}{0.15em}
DeepSeek-V3.2~\cite{deepseekai2025deepseekv32pushingfrontieropen} & 80.3 & 71.8 & 82.3 & 69.5 & 78.5 & 71.9 & 80.3 & 51.6 & 64.6 & 73.7 \\
IBA-Agent (DeepSeek-V3.2) & \uline{92.1} & \uline{82.3} & 84.8 & 79.4 & 87.7 & 79.9 & 89.2 & 57.1 & \uline{80.0} & 82.8 \\
\rowcolor{lightpurple}
IBA-Agent (DeepSeek-V3.2, FC) & \best{93.3} & \best{84.1} & \best{89.7} & \uline{81.3} & \best{90.6} & 80.4 & \best{93.3} & \uline{59.0} & \best{81.1} & \best{85.1} \\

\midrule
\specialrule{0.08em}{0.15em}{0.15em}
\multicolumn{11}{c}{\textbf{\textit{$\bullet$ Qwen3-4B-Instruct Variants}}} \\
\specialrule{0.08em}{0.15em}{0.15em}
Qwen3-4B-Instruct~\cite{qwen3technicalreport} & 79.7 & 72.3 & 82.4 & 70.5 & 80.1 & 76.1 & 84.4 & 49.5 & 64.9 & 74.6 \\
IBA-Agent (Qwen3-4B-Instruct) & 86.8 & 79.6 & 85.2 & 79.0 & 88.0 & \best{84.1} & 90.1 & 53.0 & 76.1 & 81.6 \\
\rowcolor{lightpurple}
IBA-Agent (Qwen3-4B-Instruct, FC) & \uline{88.8} & \uline{80.7} & \uline{85.3} & \best{82.2} & \uline{90.5} & \uline{83.4} & \uline{93.2} & 48.4 & \uline{76.9} & \uline{82.9} \\

\midrule
\specialrule{0.08em}{0.15em}{0.15em}
\multicolumn{11}{c}{\textbf{\textit{$\bullet$ General-Purpose AI Agents}}} \\
\specialrule{0.08em}{0.15em}{0.15em}
nanobot (DeepSeek-V3.2) & 81.5 & 64.6 & 63.2 & 68.3 & 62.0 & 54.2 & 61.8 & 41.0 & 70.5 & 64.9 \\
Claude Code (DeepSeek-V3.2) & 84.8 & 67.2 & 65.6 & 73.8 & 68.6 & 54.5 & 60.7 & 35.9 & 56.2 & 66.1 \\
Hermes Agent (DeepSeek-V3.2) & 83.8 & 67.8 & 66.1 & 73.3 & 65.6 & 56.7 & 62.8 & 37.8 & 69.0 & 67.4 \\
\bottomrule
\end{tabular}
}
\caption{Performance comparison using the \textbf{bge-m3} retriever and \textbf{Qwen3-4B-Instruct} as the judge model. Reported results are averaged over five runs. The columns correspond to nine application domains: \textbf{Writing} (writing and design), \textbf{Work}, \textbf{Daily} (daily consumption), \textbf{Plan.} (planning), \textbf{Health} (exercise and health), \textbf{Trans.} (transportation), \textbf{Med.} (medical services), \textbf{Leis.} (leisure activities), and \textbf{Info.} (information management). The best result in each column is highlighted as \colorbox{bestgreen}{\textbf{green bold text}}, the second-best result is \uline{underlined}, \colorbox{lightpurple}{Function Call variants} are shaded in light purple. FC stands for Function Call.}
\label{tab:avg_results}
\end{table*}

\begin{table*}[t]
\centering
\footnotesize
\setlength{\tabcolsep}{4.8pt}
\renewcommand{\arraystretch}{1.10}

\resizebox{\linewidth}{!}{
\begin{tabular}{>{\raggedright\arraybackslash}p{4.15cm}cccccccccc}
\toprule
\rowcolor{headergray}
\textbf{Model} & \textbf{Writing} & \textbf{Work} & \textbf{Daily} & \textbf{Plan.} & \textbf{Health} & \textbf{Trans.} & \textbf{Med.} & \textbf{Leis.} & \textbf{Info.} & \cellcolor{lowgreen}\textbf{Overall} \\
\midrule

Qwen2.5-7B-Instruct
& 66.0\std{0.57} & 61.5\std{0.30} & 79.8\std{0.47} & 64.0\std{0.29} & 75.2\std{0.15} & 66.5\std{0.34} & 78.2\std{0.14} & 47.3\std{0.62} & 45.3\std{0.53} & \overall{66.0}{0.22} \\

DeepSeek-V3
& \cellcolor{highred}83.1\std{9.48} & 69.3\std{0.42} & 78.8\std{0.23} & 70.8\std{0.28} & 78.8\std{0.38} & 68.1\std{0.17} & 81.4\std{0.21} & 49.2\std{0.42} & 64.4\std{0.11} & \overall{73.1}{1.28} \\

ChatGLM-4-9b-chat
& 66.6\std{0.67} & 57.6\std{0.37} & 74.0\std{0.55} & 62.3\std{0.18} & 73.4\std{0.48} & 60.0\std{0.16} & 71.7\std{0.31} & 42.3\std{0.83} & 46.5\std{0.83} & \overall{63.1}{0.19} \\

QwQ-32B
& \cellcolor{highred}84.4\std{8.74} & 72.6\std{0.40} & 83.4\std{0.45} & 70.4\std{0.18} & 80.6\std{0.16} & 74.4\std{0.36} & 82.1\std{0.15} & 49.3\std{0.25} & 67.0\std{0.19} & \overall{75.3}{1.06} \\

GPT-4o-mini
& 71.3\std{0.65} & 66.4\std{0.30} & 79.9\std{0.43} & 65.2\std{0.30} & 75.6\std{0.97} & 69.2\std{0.24} & 74.3\std{0.18} & 43.7\std{1.08} & 53.7\std{0.48} & \overall{68.4}{0.29} \\

GPT-4o
& 76.5\std{0.61} & 67.4\std{0.30} & 78.0\std{0.32} & 67.2\std{0.18} & 77.3\std{0.52} & 67.0\std{0.04} & 75.9\std{0.26} & 47.4\std{0.23} & 56.7\std{0.23} & \overall{69.7}{0.09} \\

GPT-5-mini
& \cellcolor{highred}81.02\std{9.76} & \cellcolor{highred}73.34\std{7.76} & 79.20\std{0.66} & \cellcolor{highred}75.16\std{5.07} & \cellcolor{midyellow}82.12\std{4.15} & \cellcolor{midyellow}72.32\std{2.09} & 82.26\std{0.50} & \cellcolor{midyellow}54.78\std{3.34} & \cellcolor{highred}73.58\std{11.01} & \overall{76.10}{4.64} \\

GPT-5.1
& 84.6\std{0.61} & 75.0\std{0.07} & 83.2\std{0.30} & 72.5\std{0.54} & 79.7\std{0.15} & 72.2\std{0.35} & 79.5\std{0.07} & 63.3\std{0.67} & 71.9\std{0.90} & \overall{76.7}{0.11} \\

\midrule
\rowcolor{groupgray}
\multicolumn{11}{l}{\textbf{DeepSeek-V3.2 Variants}} \\

DeepSeek-V3.2
& 80.3\std{0.68} & 71.8\std{0.33} & 82.3\std{0.56} & 69.5\std{0.19} & 78.5\std{0.61} & 71.9\std{0.26} & 80.3\std{0.35} & 51.6\std{1.04} & 64.6\std{0.52} & \overall{73.7}{0.24} \\

IBA-Agent (DeepSeek-V3.2)
& 92.1\std{0.88} & 82.3\std{0.34} & \cellcolor{midyellow}84.8\std{3.20} & 79.4\std{0.13} & 87.7\std{0.28} & \cellcolor{highred}79.9\std{5.48} & 89.2\std{0.70} & \cellcolor{highred}57.1\std{6.47} & 80.0\std{0.63} & \overall{82.8}{0.31} \\

IBA-Agent (DeepSeek-V3.2, FC)
& 93.3\std{0.56} & 84.1\std{0.31} & 89.7\std{0.38} & 81.3\std{0.38} & 90.6\std{0.23} & 80.4\std{0.38} & 93.3\std{0.13} & 59.0\std{0.33} & 81.1\std{0.96} & \overall{85.1}{0.23} \\

\midrule
\rowcolor{groupgray}
\multicolumn{11}{l}{\textbf{Qwen3-4B-Instruct Variants}} \\

Qwen3-4B-Instruct
& 79.7\std{0.38} & 72.3\std{0.36} & 82.4\std{0.41} & 70.5\std{0.22} & 80.1\std{0.28} & 76.1\std{0.55} & 84.4\std{0.81} & 49.5\std{0.61} & 64.9\std{0.47} & \overall{74.6}{0.15} \\

IBA-Agent (Qwen3-4B-Instruct)
& 86.8\std{0.30} & 79.6\std{0.43} & \cellcolor{midyellow}85.2\std{4.75} & 79.0\std{0.27} & 88.0\std{0.13} & \cellcolor{midyellow}84.1\std{3.14} & 90.1\std{0.18} & \cellcolor{highred}53.0\std{7.76} & 76.1\std{0.57} & \overall{81.6}{0.24} \\

IBA-Agent (Qwen3-4B-Instruct, FC)
& 88.82\std{0.78} & 80.74\std{0.35} & 85.34\std{0.43} & 82.22\std{0.30} & 90.54\std{0.26} & 83.46\std{0.23} & 93.20\std{0.16} & 48.46\std{0.78} & 76.96\std{0.69} & \overall{82.90}{0.29} \\

\midrule
\rowcolor{groupgray}
\multicolumn{11}{l}{\textbf{General-Purpose AI Agents}} \\

nanobot & 81.5\std{0.4} & 64.6\std{0.7} & 63.2\std{1.1} & 68.3\std{0.3} & 62.0\std{0.2} & 54.2\std{0.5} & 61.8\std{0.2} & 41.0\std{0.6} & 70.5\std{1.1} & \overall{64.9}{0.2} \\

Hermes Agent
& 83.8\std{0.3} & 67.8\std{0.5} & 66.1\std{0.2} & 73.3\std{0.4} & 65.6\std{0.5} & 56.7\std{0.2} & 62.8\std{0.6} & 37.8\std{0.7} & 69.0\std{0.3} & \overall{67.4}{0.1} \\

Claude Code
& \cellcolor{midyellow}84.8\std{4.7} & 67.2\std{0.5} & 65.6\std{0.6} & 73.8\std{0.3} & 68.6\std{0.3} & 54.5\std{0.4} & 60.7\std{0.2} & 35.9\std{0.6} & 56.2\std{0.5} & \overall{66.1}{0.5} \\

\bottomrule
\end{tabular}
}
\caption{Performance comparison using the \textbf{bge-m3} retriever and \textbf{Qwen3-4B-Instruct} as the judge model. Reported results are averaged over five runs, with sample standard deviations shown in a compact \(\pm\) format. The columns correspond to nine application domains: \textbf{Writing} (writing and design), \textbf{Work}, \textbf{Daily} (daily consumption), \textbf{Plan.} (planning), \textbf{Health} (exercise and health), \textbf{Trans.} (transportation), \textbf{Med.} (medical services), \textbf{Leis.} (leisure activities), and \textbf{Info.} (information management). \colorbox{lowgreen}{\strut Green} highlights the \textbf{Overall} column, \colorbox{midyellow}{\strut yellow} marks cells with moderate variation (\(\mathrm{std}\geq2.0\) and \(<5.0\)), and \colorbox{highred}{\strut red} marks cells with high variation (\(\mathrm{std}\geq5.0\)). FC stands for Function Call.}
\label{tab:stability}
\end{table*}

\begin{table*}[t]
    \centering
    \footnotesize
    \resizebox{1.0\linewidth}{!}{

   \begin{tabular}{l *{10}{>{\centering\arraybackslash}p{0.9cm}}}
        \toprule
        \rowcolor{gray!15}
        \textbf{Model} & \textbf{Writing} & \textbf{Work} & \textbf{Daily} & \textbf{Plan.} & \textbf{Health} & \textbf{Trans.} & \textbf{Medical} & \textbf{Leisure} & \textbf{Info.} & \textbf{Overall} \\
        \midrule
       
        Qwen2.5-7B-Instruct & 46.9 & 45.6 & 76.6 & 53.2 & 66.6 & 85.7 & 79.3 & 51.3 & 37.1 & 59.2 \\
        DeepSeek-V3 & 54.2 & 48.3 & 75.9 & 66.1 & 72.5 & 89.7 & 87.3 & 53.4 & 43.1 & 64.7 \\
        GPT-4o-mini & 49.9 & 45.9 & 79.0 & 54.9 & 70.0 & 92.8 & 90.5 & 58.8 & 37.6 & 62.4 \\
        GPT-4o & 49.8 & 45.4 & 77.4 & 54.3 & 66.6 & 91.8 & 87.8 & 52.0 & 35.8 & 60.8 \\
       
        \bottomrule
    \end{tabular}
    }
    \caption{Performance comparison using \textbf{all-mpnet-base-v2} retriever. The columns correspond to nine application domains: \textbf{Writing} (writing and design), \textbf{Work}, \textbf{Daily} (daily consumption), \textbf{Plan.} (planning), \textbf{Health} (exercise and health), \textbf{Trans.} (transportation), \textbf{Med.} (medical services), \textbf{Leis.} (leisure activities), and \textbf{Info.} (information management).  
    }
    \label{tab:res_mpnet}
\end{table*}

\begin{figure*}[t]
    \centering
    \includegraphics[width=\textwidth]{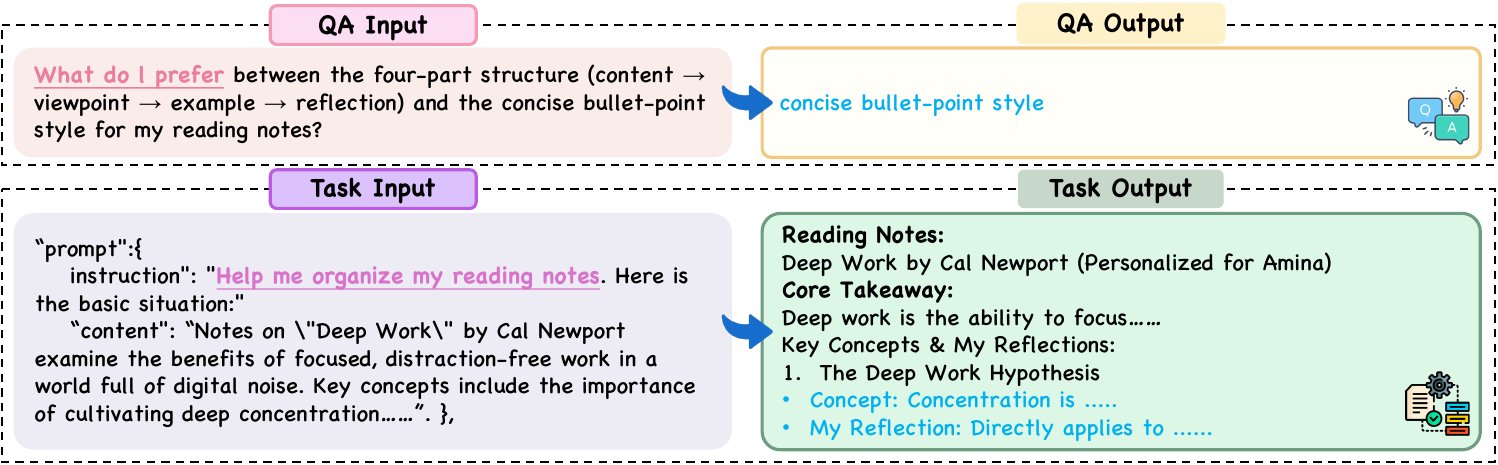}
    \caption{QA (using HiCUPID~\cite{mok25etp} as an example) vs. IBA-Agent for preference modeling. All comparisons are conducted using DeepSeek-V3.2 to control for model capability.}
    \vspace{-4mm}
    \label{fig:case_study_2}
\end{figure*}

\begin{figure}[t]
    \centering
    \includegraphics[width=0.5\textwidth]{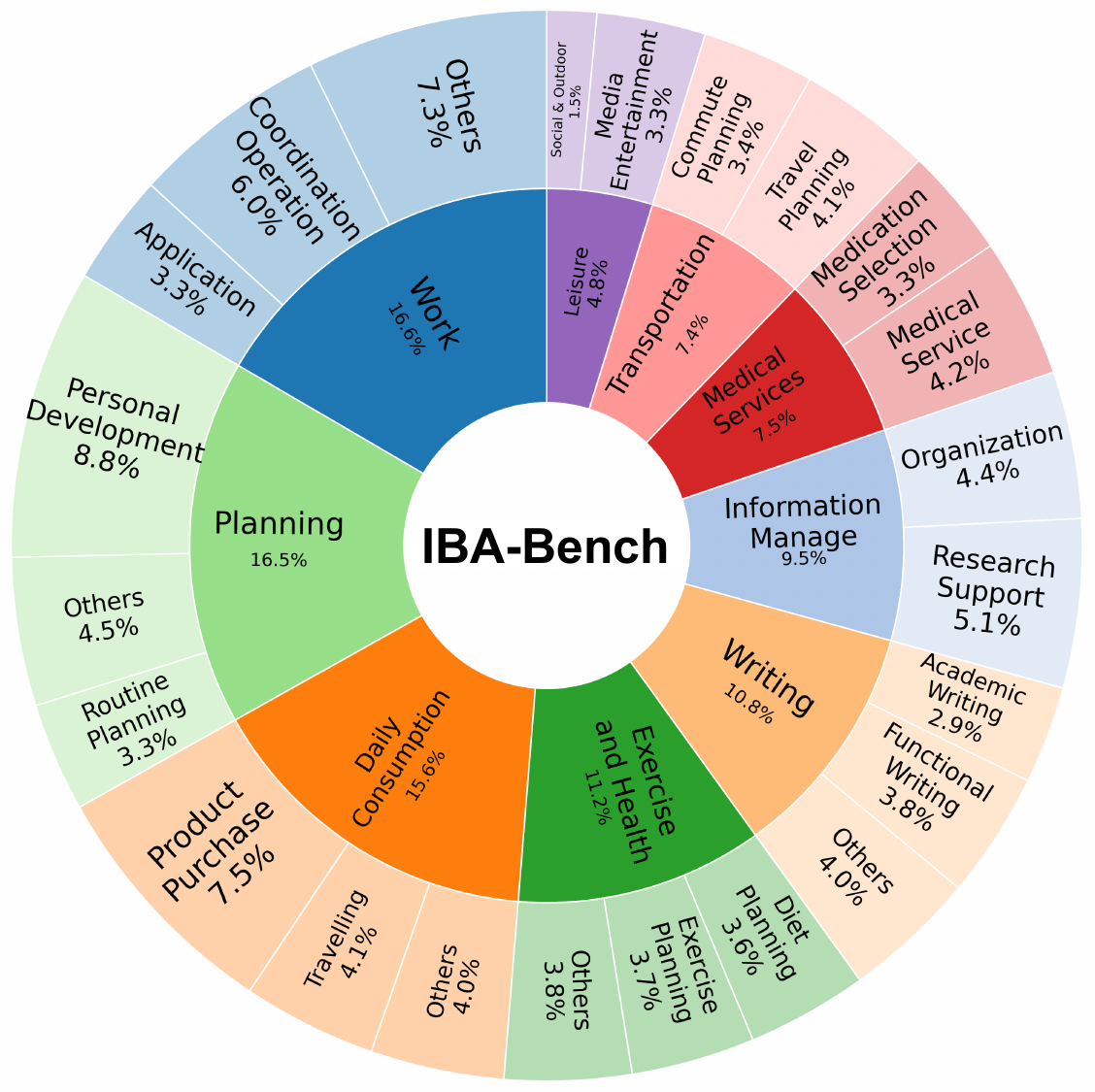}
    \vspace{-6mm}
    \caption{Hierarchical distribution of domains and scenarios in the benchmark.}

    \label{fig:data_cal}
\end{figure}

\begin{table}[t]
\centering

\small
\setlength{\tabcolsep}{12pt}
\renewcommand{\arraystretch}{1.08}
\begin{tabular}{lcc}
\toprule
\textbf{Dimension} & \textbf{Agreement} & \textbf{$AC_1$} \\
\midrule

\rowcolor{gray!15}
\multicolumn{3}{l}{\textbf{Human--Human}} \\
Reasonableness      & 73.2\% & 0.61 \\
Coherence           & 82.5\% & 0.79 \\
Persona Consistency & 85.1\% & 0.72 \\
\cmidrule(lr){1-3}
\rowcolor{gray!8}
\textbf{Overall}    & \textbf{80.2\%} & \textbf{0.70} \\

\midrule

\rowcolor{gray!15}
\multicolumn{3}{l}{\textbf{AI--AI (GPT-5.4)}} \\
Reasonableness      & 94.5\% & 0.94 \\
Coherence           & 94.5\% & 0.94 \\
Persona Consistency & 93.0\% & 0.92 \\
\cmidrule(lr){1-3}
\rowcolor{gray!8}
\textbf{Overall}    & \textbf{94.00\%} & \textbf{0.93} \\

\bottomrule
\end{tabular}
\caption{Human--human and AI--AI agreement on benchmark quality annotations over 200 matched samples.}
\label{tab:annotation_agreement}
\end{table}

\begin{table}[t]
\centering
\footnotesize
\setlength{\tabcolsep}{3pt}
\renewcommand{\arraystretch}{1.05}
\begin{tabular}{p{0.56\columnwidth}c}
\toprule
\rowcolor{gray!15}
\textbf{Signal category} & \textbf{Score contribution} \\
\midrule
Explicit revisions & 3--4 \\
User complaint or strong affect & 3--4 \\
Direct preference statement & 2 \\
Implicit behavioral patterns & 1 \\
\midrule
Retention threshold & $\geq 5$ \\
\bottomrule
\end{tabular}
\caption{Scoring criteria used by the \texttt{salience re-ranking} tool for preference-aware passage re-ranking. Implicit behavioral patterns are indirect cues, such as repeated behaviors.}
\label{tab:salience_score}
\end{table}

\section{Benchmark Details}
\subsection{Dataset Details}
\label{app:dataset_details}

Table~\ref{tab:scenario_stats} summarizes the statistics across domains and scenarios, 
including the number of preference dimensions (\textbf{D}), preference values (\textbf{P}), task instances (\textbf{I}), dynamic-preference support, and action space.
This provides a quantitative view of \textsc{IBA-Bench}’s richness in cross-domain scenario coverage and preference-space diversity.
Figure~\ref{fig:data_cal} shows the domain and scenario distributions. 

Table~\ref{tab:full_preference_taxonomy} provides illustrative examples of how preference dimensions are defined in our benchmark across the Writing and Design domain. It is organized in a domain--scenario--dimension--value hierarchy, and lists mutually exclusive preference values for each dimension to support task instantiation and controlled evaluation.

\definecolor{mygreen}{RGB}{45,150,80}
\definecolor{myred}{RGB}{200,60,60}

\definecolor{genbg}{RGB}{220,232,248}
\definecolor{gentext}{RGB}{35,63,112}
\definecolor{apibg}{RGB}{236,226,248}
\definecolor{apitext}{RGB}{112,62,145}

\newcommand{\actiontag}[3]{%
  \begingroup
  \setlength{\fboxsep}{0.25pt}%
  \colorbox{#1}{\scriptsize\textbf{\textcolor{#2}{\texttt{#3}}}}%
  \endgroup
}
\newcommand{\genaction}{\actiontag{genbg}{gentext}{Generate(text)}}
\newcommand{\apiaction}{\actiontag{apibg}{apitext}{CallAPI(params)}}
\newcommand{\groupsep}{\specialrule{0.35pt}{0.5pt}{0.5pt}}

\begin{table*}[t]
\centering 
\footnotesize
\setlength{\tabcolsep}{1.2pt} 
\renewcommand{\arraystretch}{0.95} 
\begin{tabular}{
@{}
>{\raggedright\arraybackslash}p{0.125\textwidth}
>{\raggedright\arraybackslash}p{0.310\textwidth}
>{\centering\arraybackslash}p{0.088\textwidth}
>{\centering\arraybackslash}p{0.083\textwidth}
>{\centering\arraybackslash}p{0.078\textwidth}
>{\centering\arraybackslash}p{0.065\textwidth}
>{\centering\arraybackslash}p{0.160\textwidth}
@{}
}
\toprule
\rowcolor{gray!15}
\multicolumn{1}{c}{\textbf{Domain}}
& \multicolumn{1}{c}{\textbf{Scenario}}
& \multicolumn{1}{c}{\textbf{D}}
& \multicolumn{1}{c}{\textbf{P}}
& \multicolumn{1}{c}{\textbf{I}}
& \multicolumn{1}{c}{\textbf{Dyn.}}
& \textbf{Action Space} \\
\midrule
\multirow{16}{*}{\makecell[l]{Writing and\\Design}}
 & Peer review reports & 4 & 9  & 54 & \textcolor{mygreen}{\ding{51}} & \genaction \\
 & Academic paper writing & 6 & 15 & 71 & \textcolor{mygreen}{\ding{51}} & \genaction \\
 & Research proposal (RP) writing & 4 & 13 & 76 & \textcolor{mygreen}{\ding{51}} & \genaction \\
 & Business plan writing & 3 & 8 & 29 & \textcolor{mygreen}{\ding{51}} & \genaction \\
 & Diary writing & 3 & 7 & 23 & \textcolor{mygreen}{\ding{51}} & \genaction \\
 & Technical documentation writing & 4 & 12 & 80 & \textcolor{mygreen}{\ding{51}} & \genaction \\
 & Blog post writing & 2 & 4  & 4  & \textcolor{myred}{\ding{55}} & \genaction \\
 & WeChat public account article & 3 & 8  & 27 & \textcolor{mygreen}{\ding{51}} & \genaction \\
 & Novel writing & 2 & 4  & 6 & \textcolor{myred}{\ding{55}} & \genaction \\
 & Resume writing & 5 & 14 & 72 & \textcolor{mygreen}{\ding{51}} & \genaction \\
 & Short video script design & 3 & 11 & 41  & \textcolor{myred}{\ding{55}} & \genaction \\
 & PPT slide content writing & 2 & 9 & 15  & \textcolor{mygreen}{\ding{51}} & \genaction \\
 & Mind map design & 4 & 20 & 79 & \textcolor{mygreen}{\ding{51}} & \genaction \\
 & Modular course writing (teaching aid) & 4 & 14 & 71 & \textcolor{mygreen}{\ding{51}} & \genaction \\
 & Instruction manual writing & 3 & 15 & 73 & \textcolor{mygreen}{\ding{51}} & \genaction \\
 & Self-introduction script writing & 3 & 8  & 29  & \textcolor{myred}{\ding{55}} & \genaction \\
\groupsep
\multirow{12}{*}{Work}
 & Scholarship application & 7 & 17 & 140 & \textcolor{mygreen}{\ding{51}} & \genaction \\
 & Performance report spreadsheet & 6 & 23 & 90 & \textcolor{mygreen}{\ding{51}} & \genaction \\
 & Schedule planning & 6 & 31 & 119  & \textcolor{mygreen}{\ding{51}} & \genaction \\
 & Data analysis & 4 & 27 & 101  & \textcolor{mygreen}{\ding{51}} & \genaction \\
 & Programming & 5 & 23 & 90  & \textcolor{mygreen}{\ding{51}} & \genaction \\
 & Leave application & 4 & 10 & 93  & \textcolor{myred}{\ding{55}} & \genaction \\
 & Project management & 5 & 22 & 105  & \textcolor{mygreen}{\ding{51}} & \genaction \\
 & Communication management & 2 & 5  & 7   & \textcolor{myred}{\ding{55}} & \genaction \\
 & Announcement drafting & 3 & 12 & 74  & \textcolor{mygreen}{\ding{51}} & \genaction \\
 & Event organization & 4 & 14 & 111  & \textcolor{myred}{\ding{55}} & \genaction \\
 & Work summary writing & 4 & 16  & 113 & \textcolor{mygreen}{\ding{51}} & \genaction \\
 & Terminology explanation & 3 & 14 & 112  & \textcolor{mygreen}{\ding{51}} & \genaction \\
\groupsep
\multirow{8}{*}{\makecell[l]{Daily\\Consumption}}
 & Food ordering & 3 & 15 & 122 & \textcolor{mygreen}{\ding{51}} & \apiaction \\
 & Ticket booking & 4 & 17 & 145  & \textcolor{mygreen}{\ding{51}} & \apiaction \\
 & Clothing purchase & 4 & 15 & 153  & \textcolor{mygreen}{\ding{51}} & \apiaction \\
 & Digital product purchase & 4 & 10 & 89  & \textcolor{myred}{\ding{55}} & \apiaction \\
 & Accommodation booking & 4 & 12 & 138  & \textcolor{myred}{\ding{55}} & \apiaction \\
 & Online course purchase & 5 & 12 & 154 & \textcolor{myred}{\ding{55}} & \apiaction \\
 & Book purchase & 4 & 12 & 127 & \textcolor{mygreen}{\ding{51}} & \apiaction \\
 & Renting a house & 11 & 28 & 158 & \textcolor{myred}{\ding{55}} & \apiaction \\
\groupsep
\multirow{10}{*}{Planning}
 & Weekly plan & 4 & 19 & 122 & \textcolor{mygreen}{\ding{51}} & \genaction \\
 & Monthly plan & 4 & 12 & 107 & \textcolor{mygreen}{\ding{51}} & \genaction \\
 & Long-term goal setting & 3 & 16 & 130 & \textcolor{mygreen}{\ding{51}} & \genaction \\
 & Project plan & 4 & 15 & 114  & \textcolor{myred}{\ding{55}} & \genaction \\
 & Financial plan & 3 & 13 & 74  & \textcolor{mygreen}{\ding{51}} & \genaction \\
 & Travel plan & 5 & 18 & 123  & \textcolor{mygreen}{\ding{51}} & \genaction \\
 & Study plan & 4 & 19 & 154  & \textcolor{mygreen}{\ding{51}} & \genaction \\
 & Career development plan & 4 & 16 & 100  & \textcolor{mygreen}{\ding{51}} & \genaction \\
 & Habit-building plan & 4 & 18 & 115  & \textcolor{myred}{\ding{55}} & \genaction \\
 & Skill improvement plan & 4 & 16 & 112  & \textcolor{mygreen}{\ding{51}} & \genaction \\
\groupsep
\multirow{4}{*}{\makecell[l]{Exercise and\\Health}}
 & Exercise plan & 5 & 22 & 260  & \textcolor{mygreen}{\ding{51}} & \genaction \\
 & Diet plan & 4 & 21 & 255 & \textcolor{mygreen}{\ding{51}} & \genaction \\
 & Sleep plan & 6 & 32 & 262  & \textcolor{mygreen}{\ding{51}} & \genaction \\
 & Mental health support & 1 & 6 & 4   & \textcolor{myred}{\ding{55}} & \genaction \\
\groupsep
\multirow{4}{*}{Transportation}
 & Commute route planning & 5 & 21 & 236  & \textcolor{myred}{\ding{55}} & \apiaction \\
 & Taxi / ride-hailing travel & 2 & 6  & 8   & \textcolor{myred}{\ding{55}} & \apiaction \\
 & Travel and sightseeing route planning & 4 & 19 & 215  & \textcolor{myred}{\ding{55}} & \apiaction \\
 & Parking & 3 & 10 & 59   & \textcolor{myred}{\ding{55}} & \apiaction \\
\groupsep
\multirow{3}{*}{\makecell[l]{Medical\\Services}}
 & Appointment registration & 4 & 15 & 210 & \textcolor{myred}{\ding{55}} & \apiaction \\
 & Medicine selection & 4 & 17 & 231  & \textcolor{myred}{\ding{55}} & \apiaction \\
 & Online medical consultation & 3 & 14 & 83  & \textcolor{myred}{\ding{55}} & \genaction \\
\groupsep
\multirow{3}{*}{\makecell[l]{Leisure\\Activities}}
 & Audio-visual entertainment & 4 & 24 & 229  & \textcolor{mygreen}{\ding{51}} & \apiaction \\
 & Outdoor leisure & 2 & 5  & 6  & \textcolor{myred}{\ding{55}} & \apiaction \\
 & Social assistance & 3 & 14 & 98  & \textcolor{mygreen}{\ding{51}} & \genaction \\
\groupsep
\multirow{6}{*}{\makecell[l]{Information\\Management}}
 & Research report writing & 4 & 18 & 174 & \textcolor{mygreen}{\ding{51}} & \genaction \\
 & Timeline organization & 3 & 10 & 41  & \textcolor{mygreen}{\ding{51}} & \genaction \\
 & Meeting minutes organization & 3 & 13 & 60  & \textcolor{mygreen}{\ding{51}} & \genaction \\
 & Reading notes & 2 & 9  & 16  & \textcolor{mygreen}{\ding{51}} & \genaction \\
 & Folder organization & 5 & 19 & 191  & \textcolor{myred}{\ding{55}} & \genaction \\
 & Tool usage & 9 & 34 & 182 & \textcolor{mygreen}{\ding{51}} & \genaction \\
\bottomrule
\end{tabular}
\caption{
Scenario-level statistics across benchmark domains.
\textbf{D}: number of preference dimensions;
\textbf{P}: number of preference values defined for the scenario;
\textbf{I}: number of task instances;
\textbf{Dyn.}: whether dynamic preferences are supported.
}
\label{tab:scenario_stats}
\end{table*}

\begin{table*}[t]
\centering
\scriptsize
\renewcommand{\arraystretch}{1.05}
\setlength{\tabcolsep}{3pt}

\begin{tabular}{
p{0.15\textwidth}
p{0.17\textwidth}
p{0.17\textwidth}
p{0.40\textwidth}
}

\toprule
\rowcolor{gray!15}
\multicolumn{1}{l}{\textbf{Domain}} &
\multicolumn{1}{l}{\textbf{Scenario}} &
\multicolumn{1}{l}{\textbf{Preference Dimension}} &
\multicolumn{1}{l}{\textbf{Preference Value}} \\
\midrule

\multirow[c]{7}{0.15\textwidth}{Writing and Design}
& \multirow[c]{7}{0.17\textwidth}{Journal writing}
& \multirow[c]{2}{0.17\textwidth}{Time granularity}
& Records by day (one entry per whole day). \\
& & & Records by key moments/timestamps. \\
\cmidrule(lr){3-4}

& & \multirow[c]{3}{0.17\textwidth}{Content structure}
& Prefers a fixed template: ``What happened / How I felt''. \\
& & & Prefers free-form writing (write whatever comes to mind). \\
& & & Preferred the fixed template; now prefers free-form writing. \\
\cmidrule(lr){3-4}

& & \multirow[c]{2}{0.17\textwidth}{Length preference}
& Prefers short entries (about 50--150 words each). \\
& & & Prefers long entries (detailed expansion; 300+ words). \\

\midrule

\multirow[c]{8}{0.15\textwidth}{Writing and Design}
& \multirow[c]{8}{0.17\textwidth}{Business plan writing}
& \multirow[c]{4}{0.17\textwidth}{Overall narrative style}
& Prefers narrative-driven writing. \\
& & & Prefers logic-driven writing (clear structure, data-centric). \\
& & & Preferred logic-driven writing; now narrative-driven writing. \\
& & & Preferred narrative-driven writing; now logic-driven writing. \\
\cmidrule(lr){3-4}

& & \multirow[c]{2}{0.17\textwidth}{Market analysis depth}
& Prefers extensive use of industry data. \\
& & & Prefers highlighting key data points with concise interpretation. \\
\cmidrule(lr){3-4}

& & \multirow[c]{2}{0.17\textwidth}{Risk section writing}
& Emphasizes explicit identification of risks. \\
& & & Follows an optimistic narrative, downplays risks. \\

\midrule

\multirow[c]{12}{0.15\textwidth}{Writing and Design}
& \multirow[c]{12}{0.17\textwidth}{Technical doc writing}
& \multirow[c]{2}{0.17\textwidth}{Document structure}
& Task-oriented structure organized around ``how-to'' steps. \\
& & & Concept-oriented structure. \\
\cmidrule(lr){3-4}

& & \multirow[c]{2}{0.17\textwidth}{Technical depth}
& Prefers detailed explanations. \\
& & & Prefers simplified explanations. \\
\cmidrule(lr){3-4}

& & \multirow[c]{6}{0.17\textwidth}{Example code style}
& Prefers complete, runnable code examples. \\
& & & Prefers code snippets (key parts only). \\
& & & Prefers pseudocode. \\
& & & Previously preferred complete; now prefers pseudocode. \\
& & & Previously preferred pseudocode; now prefers code snippets. \\
& & & Previously preferred code snippets; now prefers pseudocode. \\
\cmidrule(lr){3-4}

& & \multirow[c]{2}{0.17\textwidth}{API explanation}
& Prefers table-style specification. \\
& & & Prefers narrative explanation of API behavior. \\

\midrule

\multirow[c]{9}{0.15\textwidth}{Writing and Design}
& \multirow[c]{9}{0.17\textwidth}{Review writing}
& \multirow[c]{2}{0.17\textwidth}{Overall structure}
& Prefers organizing reviews into Summary / Pros / Cons sections. \\
& & & Prefers unstructured summaries with weaknesses separated. \\
\cmidrule(lr){3-4}

& & \multirow[c]{2}{0.17\textwidth}{Weakness format}
& Prefers bullet-pointed weaknesses (with numbering). \\
& & & Prefers bullet-pointed weaknesses (without numbering). \\
\cmidrule(lr){3-4}

& & \multirow[c]{2}{0.17\textwidth}{Weakness count}
& Requires exactly three weaknesses. \\
& & & Requires exactly five weaknesses. \\
\cmidrule(lr){3-4}

& & \multirow[c]{3}{0.17\textwidth}{Weakness content}
& Prefers keywords + full sentences. \\
& & & Prefers direct full-sentence descriptions. \\
& & & Preferred direct sentences; now prefers keywords + sentences. \\

\midrule

\multirow[c]{4}{0.15\textwidth}{Writing and Design}
& \multirow[c]{4}{0.17\textwidth}{Blog article writing}
& \multirow[c]{2}{0.17\textwidth}{Narrative perspective}
& First-person personal perspective. \\
& & & Third-person objective perspective. \\
\cmidrule(lr){3-4}

& & \multirow[c]{2}{0.17\textwidth}{Citations / sources}
& Prefers academic style: formal citations, links, references. \\
& & & Prefers original style. \\

\midrule

\multirow[c]{8}{0.15\textwidth}{Writing and Design}
& \multirow[c]{8}{0.17\textwidth}{Public account writing}
& \multirow[c]{4}{0.17\textwidth}{Opening style}
& Storytelling-based opening. \\
& & & Viewpoint-based opening. \\
& & & Preferred storytelling-based opening, now viewpoint-based. \\
& & & Preferred viewpoint-based opening, now storytelling-based. \\
\cmidrule(lr){3-4}

& & \multirow[c]{2}{0.17\textwidth}{Social expression}
& Leans toward social topics. \\
& & & Leans toward personal perspectives. \\
\cmidrule(lr){3-4}

& & \multirow[c]{2}{0.17\textwidth}{Title style}
& Attention-grabbing titles. \\
& & & Formal titles (standard phrasing, fewer stylistic markers). \\

\midrule

\multirow[c]{4}{0.15\textwidth}{Writing and Design}
& \multirow[c]{4}{0.17\textwidth}{Novel writing}
& \multirow[c]{2}{0.17\textwidth}{Narrative perspective}
& First-person narration. \\
& & & Third-person narration. \\
\cmidrule(lr){3-4}

& & \multirow[c]{2}{0.17\textwidth}{Narrative type}
& Linear storytelling in chronological order. \\
& & & Non-linear storytelling (e.g., flashbacks, parallel storylines). \\

\bottomrule
\end{tabular}
\caption{Examples of preference dimensions and values in the Writing and Design domain.}
\label{tab:full_preference_taxonomy}
\end{table*}

\subsection{Task Formulation and Evaluation Details}
\label{app:task_define}
This section provides the detailed evaluation protocol for the three final-action types in $\mathcal{Y}$. 
For all types, the evaluator focuses on the final action rather than whether the agent can merely describe or recall user preferences. Table~\ref{tab:scenario_stats} reports the action type of each scenario, including 50 \texttt{Generate(text)} and 16 \texttt{CallAPI(params)} scenarios.

\noindent\textbf{\texttt{Generate(text)}.} 
This type is used for text-oriented tasks where the agent completes the request by producing a free-form personalized response, such as a plan, explanation, or written message. 
Each instance contains a set of hidden personalized requirements for a specific user, such as style preferences, structural preferences, or situational needs. 
The evaluator assigns a binary score to each requirement according to whether the generated response satisfies it, and the instance score is the average over all requirements. 
An example is shown in Table~\ref{tab:generate_example}.

\noindent\textbf{\texttt{CallAPI(params)}.  }
This type evaluates whether an agent can translate preferences into executable tool parameters. 
In scenarios such as ordering food or booking appointments, the agent's final action is represented as structured API parameters that can be passed to the corresponding tool interface to perform the intended operation. 
Each instance defines an API template with controllable slots, and the hidden personalized requirements specify which slots should reflect the user's preferences and current situation. 
The evaluator scores these personalization-relevant slots independently: each slot receives a binary score indicating whether the submitted value satisfies the user's preferences in the given scenario, and the instance score is the average over all scored slots.
Task-fixed or API-validity-only parameters are excluded from the personalization score.
An example is shown in Table~\ref{tab:api_example}.

\begin{table*}[t]
\centering
\scriptsize
\setlength{\tabcolsep}{4pt}
\renewcommand{\arraystretch}{1.12}

\definecolor{exblue}{RGB}{38,70,145}
\definecolor{exgreen}{RGB}{42,145,62}
\definecolor{exred}{RGB}{185,55,55}
\definecolor{exgray}{RGB}{245,245,245}
\definecolor{exdark}{RGB}{64,128,210}
\definecolor{apidark}{RGB}{64,128,210}

\begin{tcolorbox}[
    enhanced,
    width=0.98\textwidth,
    colback=gray!2,
    colframe=gray!45,
    boxrule=0.7pt,
    arc=2pt,
    left=7pt,
    right=7pt,
    top=6pt,
    bottom=6pt,
    title={\textsc{Example 1: \texttt{Generate(text)}}},
    coltitle=white,
    colbacktitle=exdark,
    fonttitle=\bfseries,
    boxed title style={
        sharp corners,
        arc=1pt,
        boxrule=0pt,
        left=6pt,
        right=6pt,
        top=4.0pt,
        bottom=3.5pt,
        colback=exdark
    }
]

\noindent
\begin{tabularx}{\linewidth}{>{\raggedright\arraybackslash}p{0.21\linewidth}
                                >{\raggedright\arraybackslash}p{0.20\linewidth}
                                >{\raggedright\arraybackslash}X}
\toprule
\textbf{\textcolor{exblue}{Output type}} 
& \textbf{\textcolor{exblue}{Scenario}} 
& \textbf{\textcolor{exblue}{Current query}} \\
\midrule
\texttt{Generate(text)} 
& Diet planning 
& \textit{Help me create a diet plan. } \\
\bottomrule
\end{tabularx}

\vspace{0.45\baselineskip}

\noindent
\colorbox{exgray}{
\begin{minipage}{0.965\textwidth}
\vspace{0.25\baselineskip}
\noindent\textbf{\textcolor{exblue}{Hidden evaluation instance.}}
The generated plan should satisfy three personalized requirements:
\textbf{(1)} muscle gain with about $+300$ kcal/day;
\textbf{(2)} low fat, with fat intake $\leq 20\%$ of total calories;
\textbf{(3)} breakfast is mandatory.
\vspace{0.25\baselineskip}
\end{minipage}
}

\vspace{0.55\baselineskip}

\noindent
\textbf{\textcolor{exblue}{Agent output (summary version).}}
\vspace{0.15\baselineskip}

\begin{tcolorbox}[
    width=\linewidth,
    colback=white,
    colframe=gray!35,
    boxrule=0.5pt,
    arc=1.5pt,
    left=5pt,
    right=5pt,
    top=4pt,
    bottom=4pt
]
\emph{
I will make breakfast a central, non-negotiable part of the plan so it supports your muscle gain and keeps your day structured. 
Eat about $+300$ kcal above maintenance overall, with fat kept at or below $20\%$ of total calories. 
Use a fixed daily breakfast within 60--90 minutes of waking, targeting about 400--550 kcal and 25--30 g protein. 
Breakfast options include protein oatmeal, an egg-and-toast plate, or a low-fat smoothie. 
Lunch and dinner use lean protein, rice/quinoa/potatoes, vegetables, and measured oil; dinner may include a small portion of avocado or nuts.
}
\end{tcolorbox}

\vspace{0.45\baselineskip}

\noindent
\textbf{\textcolor{exblue}{Behavioral score.}}
\vspace{0.15\baselineskip}

\begin{center}
\begin{tabularx}{0.96\textwidth}{
    >{\raggedright\arraybackslash}p{0.34\textwidth}
    >{\centering\arraybackslash}p{0.08\textwidth}
    >{\raggedright\arraybackslash}X}
\toprule
\textbf{Criterion} & \textbf{Score} & \textbf{Evaluator rationale} \\
\midrule
Muscle gain with $+300$ kcal/day
& \textcolor{exgreen}{\textbf{1}}
& The plan provides a $+300$ kcal surplus and supports muscle gain. \\

Low fat, $\leq 20\%$ of calories
& \textcolor{exred}{\textbf{0}}
& Fat sources such as seeds, avocado, nuts, and oil may exceed the fat cap. \\

Breakfast is mandatory
& \textcolor{exgreen}{\textbf{1}}
& Breakfast is explicitly included as a detailed daily meal. \\
\midrule
\textbf{Total / Average}
& \textbf{2/3}
& \textbf{Average score: 0.67} \\
\bottomrule
\end{tabularx}
\end{center}

\end{tcolorbox}

\caption{Example of a \texttt{Generate(text)} task in \textsc{IBA-Bench}. }
\label{tab:generate_example}
\end{table*}

\begin{table*}[t]
\centering
\scriptsize
\setlength{\tabcolsep}{4pt}
\renewcommand{\arraystretch}{1.12}

\definecolor{apiblue}{RGB}{38,70,145}
\definecolor{apigreen}{RGB}{42,145,62}
\definecolor{apired}{RGB}{185,55,55}
\definecolor{apigray}{RGB}{245,245,245}
\definecolor{apineutral}{RGB}{95,95,95}
\definecolor{exdark}{RGB}{64,128,210}
\definecolor{apidark}{RGB}{64,128,210}

\begin{tcolorbox}[
    enhanced,
    width=0.98\textwidth,
    colback=gray!2,
    colframe=gray!45,
    boxrule=0.7pt,
    arc=2pt,
    left=7pt,
    right=7pt,
    top=6pt,
    bottom=6pt,
    title={\textsc{Example 2: \texttt{CallAPI(params)}}},
    coltitle=white,
    colbacktitle=apidark,
    fonttitle=\bfseries,
    boxed title style={
        sharp corners,
        arc=1pt,
        boxrule=0pt,
        left=6pt,
        right=6pt,
        top=4.0pt,
        bottom=3.5pt,
        colback=apidark
    }
]

\noindent
\begin{tabularx}{\linewidth}{
    >{\raggedright\arraybackslash}p{0.20\linewidth}
    >{\raggedright\arraybackslash}p{0.20\linewidth}
    >{\raggedright\arraybackslash}X}
\toprule
\textbf{\textcolor{apiblue}{Output type}} 
& \textbf{\textcolor{apiblue}{Scenario}} 
& \textbf{\textcolor{apiblue}{Current query}} \\
\midrule
\texttt{CallAPI(params)} 
& Food delivery 
& \textit{Help me order a food delivery.} \\
\bottomrule
\end{tabularx}

\vspace{0.45\baselineskip}

\noindent
\colorbox{apigray}{
\begin{minipage}{0.965\textwidth}
\vspace{0.25\baselineskip}
\noindent\textbf{\textcolor{apiblue}{Hidden evaluation instance.}}
The API action should satisfy three personalized requirements:
\textbf{(1)} the user used to like spicy delivery food, but has recently caught a cold and cannot eat spicy food;
\textbf{(2)} the user prefers Western cuisine;
\textbf{(3)} the preferred price range is 20--30 yuan.
\vspace{0.25\baselineskip}
\end{minipage}
}

\vspace{0.55\baselineskip}

\noindent
\textbf{\textcolor{apiblue}{Agent output (summary version).}}
\vspace{0.1\baselineskip}

\begin{tcolorbox}[
    width=\linewidth,
    colback=white,
    colframe=gray!35,
    boxrule=0.5pt,
    arc=1.5pt,
    left=6pt,
    right=6pt,
    top=5pt,
    bottom=5pt
]

\begin{minipage}[t]{0.55\linewidth}
\textbf{Filled API template}

\vspace{0.1\baselineskip}
{\ttfamily\scriptsize
https://meituan.com/waimai/search?\\
\hspace*{0.8em}cuisine\_type=\textcolor{apigreen}{\textbf{Western}}\&\\
\hspace*{0.8em}meal\_type=\textcolor{apineutral}{lunch}\&\\
\hspace*{0.8em}price\_range=\textcolor{apigreen}{\textbf{under 30}}\&\\
\hspace*{0.8em}flavor\_preference=\textcolor{apired}{\textbf{mild}}\&\\
\hspace*{0.8em}dietary\_restrictions=\textcolor{apired}{\textbf{non-spicy, no seafood}}\&\\
\hspace*{0.8em}restaurant\_name=\textcolor{apineutral}{none}\&\\
\hspace*{0.8em}quantity=\textcolor{apineutral}{1}\&\\
\hspace*{0.8em}time\_specific\_preference=\textcolor{apineutral}{none}
}
\end{minipage}
\hfill
\begin{minipage}[t]{0.41\linewidth}
\textbf{Reasoning: }

\vspace{0.1\baselineskip}
The agent overrides the user's previous spicy-food preference because the user recently caught a cold. 
It therefore sets \textcolor{apired}{\textbf{mild}} and \textcolor{apired}{\textbf{non-spicy}}. 
It also preserves the user's \textcolor{apigreen}{\textbf{Western cuisine}} preference and budget by setting \textcolor{apigreen}{\textbf{under 30}}.

\vspace{1.45\baselineskip}
\textcolor{apigreen}{\rule{0.8em}{0.8em}} Long-term preference / budget

\vspace{0.15\baselineskip}
\textcolor{apired}{\rule{0.8em}{0.8em}} Dynamic health-related update

\vspace{0.15\baselineskip}
\textcolor{apineutral}{\rule{0.8em}{0.8em}} Other parameters
\end{minipage}

\end{tcolorbox}

\vspace{0.35\baselineskip}

\noindent
\textbf{\textcolor{apiblue}{Behavioral score.}}
\vspace{0.1\baselineskip}

\begin{center}
\begin{tabularx}{0.96\textwidth}{
    >{\raggedright\arraybackslash}p{0.33\textwidth}
    >{\centering\arraybackslash}p{0.07\textwidth}
    >{\raggedright\arraybackslash}X}
\toprule
\textbf{Criterion} & \textbf{Score} & \textbf{Evaluator rationale} \\
\midrule
Avoid spicy food due to recent cold
& \textcolor{apigreen}{\textbf{1}}
& Uses \texttt{mild} and \texttt{non-spicy}. \\

Prefer Western cuisine
& \textcolor{apigreen}{\textbf{1}}
& Sets \texttt{cuisine\_type=Western}. \\

Price range 20--30 yuan
& \textcolor{apigreen}{\textbf{1}}
& Sets \texttt{price\_range=under 30}. \\
\midrule
\textbf{Total / Average}
& \textbf{3/3}
& \textbf{Average score: 1.00} \\
\bottomrule
\end{tabularx}
\end{center}

\end{tcolorbox}

\caption{Example of a \texttt{CallAPI(params)} task in \textsc{IBA-Bench}. }
\label{tab:api_example}
\end{table*}

\subsection{Prompt Details}

\noindent Figure~\ref{fig:prompt_casualchat_user} shows the prompt template used to generate casual-chat dialogues in \textsc{IBA-Bench}. 
Figures~\ref{fig:prompt_user_evidence} and~\ref{fig:system_prompt_evidence} show the prompt templates used to synthesize user historical dialogue evidence.

\section{Benchmark Annotation and Validation}
\subsection{Human Annotation Protocol}
\label{app:human_validation}

To evaluate the quality of the generated benchmark instances, we randomly sampled 200 instances for manual validation. Each instance was independently annotated by two human annotators along three binary dimensions: \textit{reasonableness}, \textit{coherence}, and \textit{persona consistency}. These dimensions are designed to assess whether an instance is reasonable, internally consistent, and aligned with the intended user role and evolving preferences over time.

\paragraph{Reasonableness.}
Whether the interaction history and instantiated task are reasonable under common sense and the task setting, without obvious hallucinations or unrealistic assumptions. \textbf{1}: plausible and credible. \textbf{0}: clearly implausible or severely hallucinatory.

\paragraph{Coherence.}
Whether the interaction history and task are logically connected and internally self-consistent, without major contradictions or abrupt discontinuities. \textbf{1}: coherent and understandable. \textbf{0}: conflicting, broken, or inconsistent.

\paragraph{Persona Consistency.}
Whether the instance is consistent with the target role and appropriately reflects preference changes over time. \textbf{1}: consistent with the intended role and preference evolution. \textbf{0}: inconsistent with the role or misrepresents preference dynamics.

\paragraph{Annotation procedure.}
The two human annotators independently labeled all sampled instances using the criteria described above. In addition, we obtained two independent AI-based annotations from GPT-5.4 using the same evaluation dimensions and instructions. To assess the reliability of the validation process, we compared (1) the agreement between the two human annotators and (2) the agreement between the two AI-based annotations. Human annotations were treated as the primary reference, while AI judgments served as an additional external comparison signal.

Table~\ref{tab:annotation_agreement} reports two complementary measures of annotation reliability: raw agreement and Gwet's $AC_1$ \cite{gwet2001handbook}. Raw agreement is the proportion of instances receiving the same label from two annotators:
\[
P_o = \frac{1}{N}\sum_{i=1}^{N} \mathbf{1}(y_i^{(1)} = y_i^{(2)}),
\]
where $N$ is the number of annotated instances, $y_i^{(1)}$ and $y_i^{(2)}$ denote the labels assigned by two raters to instance $i$, and $\mathbf{1}(\cdot)$ is the indicator function. Because the labels are imbalanced, we additionally report Gwet's $AC_1$, which is more robust than Cohen's kappa under skewed label distributions:
\[
AC_1 = \frac{P_o - P_e}{1 - P_e}
\]
where $P_o$ is the observed agreement and $P_e$ is the chance agreement estimated under Gwet's formulation.

\section{Experiment Settings}
\label{sec:exp_settings}

\subsection{Baselines} 
We evaluate personalized task completion across a diverse set of LLMs, including Qwen2.5-7B-Instruct~\cite{qwen2025qwen25technicalreport}, ChatGLM-4-9B-Chat~\cite{glm2024chatglmfamilylargelanguage}, DeepSeek-V3~\cite{deepseekai2025deepseekv3technicalreport}, DeepSeek-V3.2~\cite{deepseekai2025deepseekv32pushingfrontieropen}, QwQ-32B~\cite{QwQ32BBlog2025}, GPT-4o-mini, GPT-4o~\cite{OpenAI_GPT4o2024}, GPT-5-mini, GPT-5.1~\cite{OpenAI_GPT5_1_2025}, and Qwen3-4B-Instruct~\cite{qwen3technicalreport}, under a retrieval-augmented generation (RAG) baseline. All models are accessed via official or widely used inference platforms (DeepSeek API, SiliconFlow, and OpenAI API). The RAG baseline uses bge-m3\footnotemark[1] to retrieve sentence-level chunks concatenated with the task query as LLM input. Beyond standalone LLMs, we also evaluate representative agent-based systems, namely Claude Code, Hermes Agent, and nanobot (a lightweight personal AI agent inspired by OpenClaw), to assess their end-to-end task-completion performance under the same personalized setting.

\subsection{Evaluation Metric}
For each scenario, we define a small set of behavioral checkpoints, each corresponding to an atomic seed preference of the user in that scenario. 
For generation-based tasks, given an output $y_i$ for instance $i$, we use three LLM-based judges, namely DeepSeek-V3.2, Qwen3-4B-Instruct, and Qwen3-30B-A3B-Instruct, to assess whether $y_i$ satisfies each checkpoint, yielding a binary score $f_{i,j}^k(y_i) \in \{0,1\}$.
The instance-level score is computed as the average satisfaction across checkpoints. 
For API-actionable tasks, each required parameter is evaluated as a checkpoint, receiving $1$ if it is valid and consistent with the task requirement and inferred user preference, and $0$ otherwise. 
The instance-level score is the average correctness across required parameters.

\subsection{Parameters}
We evaluate all models using three judges: DeepSeek-V3.2, Qwen3-4B-Instruct, and Qwen3-30B-A3B-Instruct, with the maximum output length set to 3,000 tokens. For Qwen3-4B-Instruct, we report the average over five runs to assess evaluation stability, with results shown in Table~\ref{tab:stability} in Appendix~A. For retrieval, we use bge-m3 with 256-token chunks and a 50-token overlap. Standard retrieval returns the top-3 passages per query, while Broad Retrieval issues up to five expanded queries and retains at most ten passages after filtering and optional re-ranking. We fix the random seed to 42 for reproducibility.

\section{Supplementary Experiments}
\label{sec:supp_experiments}
Table~\ref{tab:res_mpnet} reports supplementary benchmark results using the \texttt{all-mpnet-base-v2} retriever, evaluating Qwen2.5-7B-Instruct, DeepSeek-V3, GPT-4o-mini, and GPT-4o.
Compared with the \texttt{bge-m3} setting in Table~\ref{tab:res_bge}, \texttt{all-mpnet-base-v2} improves the overall score for all four models (+6.1 for Qwen2.5-7B, +7.6 for DeepSeek-V3, +4.4 for GPT-4o-mini, and +3.1 for GPT-4o), with the largest gains concentrated in the Planning and Health domains. 
Tables~\ref{tab:res_30b}, \ref{tab:res_ds}, and \ref{tab:avg_results} present evaluation results on the same set of execution outputs using three different judge models, namely Qwen3-30B-A3B-Instruct, DeepSeek-V3.2, and Qwen3-4B-Instruct. Among them, Table~\ref{tab:stability} further reports the per-cell standard deviation across five repeated evaluations under Qwen3-4B-Instruct as the judge. The results show that overall scores remain highly stable for most models and IBA-Agent variants, with overall standard deviations generally below 0.31, suggesting that the main performance trends are highly reproducible.
Across three judge models, IBA-Agent consistently outperforms all baseline models, and the function-calling variant further improves overall performance in most settings, demonstrating the robustness of structured history modeling and modular agent execution. 

\FloatBarrier
\begin{figure*}[!t]
\centering

\definecolor{softtitle}{RGB}{92,92,92}
\definecolor{softblue}{RGB}{38,70,145}
\definecolor{softgreen}{RGB}{42,145,62}
\definecolor{softred}{RGB}{185,55,55}
\definecolor{softpurple}{RGB}{170,75,220}
\definecolor{softgray}{RGB}{95,95,95}

\begin{tcolorbox}[
    enhanced,
    width=0.96\textwidth,
    colback=gray!2,
    colframe=gray!50,
    boxrule=0.7pt,
    arc=2pt,
    left=8pt,
    right=8pt,
    top=5pt,
    bottom=5pt,
    drop shadow={black!12!white},
    title={\textsc{System Prompt for Casual-Chat Generation}},
    coltitle=white,
    colbacktitle=softtitle,
    fonttitle=\bfseries\large,
    fontupper=\ttfamily\footnotesize,
    boxed title style={
        sharp corners,
        arc=1pt,
        boxrule=0pt,
        left=8pt,
        right=8pt,
        top=3pt,
        bottom=3pt,
        colback=softtitle
    }
]

\footnotesize
\setlength{\baselineskip}{1.04\baselineskip}

\noindent
\textbf{\textcolor{softblue}{Role Setting.}}
You are an English casual-chat dialogue generation assistant.\\
The current user’s name is “\textcolor{softpurple}{\{user\_name\}},” and their general mood/state is “\textcolor{softpurple}{\{mood\}}.” Their detailed persona is as follows:

\vspace{2pt}
\noindent
\textcolor{softpurple}{\texttt{\{persona\_summary\}}}

\vspace{4pt}
\noindent
{\color{softblue!70}\rule{\linewidth}{0.4pt}}

\vspace{4pt}
\noindent
\textbf{\textcolor{softgreen}{Task.}}
Your task is to generate natural, realistic daily casual-chat content 
\textbf{based solely on the personal information in the above user persona} 
(such as age, major, school, living environment, interests, emotions, etc.).

\vspace{4pt}
\noindent
{\color{softgreen!75}\rule{\linewidth}{0.4pt}}

\vspace{4pt}
\noindent
\textbf{\textcolor{softred}{Important requirements:}}
\begin{enumerate}[leftmargin=1.6em, itemsep=1.5pt, topsep=2pt, label=\textcolor{softred}{\textbf{\arabic*.}}]
    \item The conversation should only be about light everyday topics, such as study rhythm, life status, social life, hobbies, emotional feelings, city/campus experiences, etc.
    \item \textbf{Absolutely no specific tasks or instructions}, especially anything like “help me review a paper,” “write reviewer comments,” “this paper/manuscript/research,” “task/instruction/system prompt,” etc.
    \item Even if the user persona includes research direction, paper topics, or project background, these can only serve as background atmosphere — do not expand on research/paper content or switch into a “task-helping” mode.
    \item The conversation may include only two speaker prefixes: “User:” and “ChatGPT:”. No other notes or markers.
    \item The overall tone should be natural and unforced — more like casual chatting between friends than executing a task.
\end{enumerate}

\end{tcolorbox}

\caption{System prompt used for casual-chat generation.}
\label{fig:prompt_casualchat_user}
\end{figure*}

\begin{figure*}[!t]
\centering

\definecolor{softtitle}{RGB}{92,92,92}
\definecolor{softblue}{RGB}{38,70,145}
\definecolor{softgreen}{RGB}{42,145,62}
\definecolor{softred}{RGB}{185,55,55}
\definecolor{softpurple}{RGB}{170,75,220}
\definecolor{softgray}{RGB}{95,95,95}

\begin{tcolorbox}[
    enhanced,
    width=0.96\textwidth,
    colback=gray!2,
    colframe=gray!50,
    boxrule=0.7pt,
    arc=2pt,
    left=8pt,
    right=8pt,
    top=5pt,
    bottom=5pt,
    drop shadow={black!12!white},
    title={\textsc{User Prompt for Evidence Data Generation}},
    coltitle=white,
    colbacktitle=softtitle,
    fonttitle=\bfseries\large,
    fontupper=\ttfamily\footnotesize,
    boxed title style={
        sharp corners,
        arc=1pt,
        boxrule=0pt,
        left=8pt,
        right=8pt,
        top=3pt,
        bottom=3pt,
        colback=softtitle
    }
]

\footnotesize
\setlength{\baselineskip}{1.03\baselineskip}

\noindent
\textbf{\textcolor{softblue}{INPUTS}}
\vspace{0.12\baselineskip}

\noindent
{\color{softblue!70}\rule{\linewidth}{0.4pt}}
\vspace{0.22\baselineskip}

\noindent
\textbf{\textcolor{softgray}{User Information.}} \\
\textcolor{softpurple}{\textit{\{person\_info\}}}

\vspace{0.22\baselineskip}

\noindent
\textbf{\textcolor{softgray}{User Preferences}}
\textcolor{softred}{\textit{(Hidden — for the scriptwriter only).}} \\
\textcolor{softpurple}{\textit{\{instance\}}}

\vspace{0.22\baselineskip}

\noindent
\textbf{\textcolor{softgray}{Task Overview.}} \\
\textcolor{softpurple}{\textit{\{instruction\_text\}}}

\vspace{0.22\baselineskip}

\noindent
\textbf{\textcolor{softgray}{Domain / Category.}} \\
\textcolor{softpurple}{\textit{\{domain\}}}

\vspace{0.22\baselineskip}

\noindent
\textbf{\textcolor{softgray}{Sub-scenario.}} \\
\textcolor{softpurple}{\textit{\{sub\_scenario\}}}

\vspace{0.22\baselineskip}

\noindent
\textbf{\textcolor{softgray}{Task Context / Available Materials.}} \\
\textcolor{softpurple}{\textit{\{content\_text\}}}

\vspace{0.38\baselineskip}

\noindent
\textbf{\textcolor{softgreen}{TASK}}
\vspace{0.12\baselineskip}

\noindent
{\color{softgreen!75}\rule{\linewidth}{0.4pt}}
\vspace{0.25\baselineskip}

Generate a complete, natural, and fluent English conversation in which a user and 
\textbf{\textcolor{softblue}{``ChatGPT''}} collaboratively accomplish:

\vspace{0.18\baselineskip}

\noindent
\textcolor{softpurple}{\textit{\{scenario\}}}

\vspace{0.22\baselineskip}

\noindent
\textbf{\textcolor{softred}{Generate the full dialogue now.}}

\end{tcolorbox}

\caption{User prompt used for evidence data generation.}
\label{fig:prompt_user_evidence}
\end{figure*}

\begin{figure*}[!t]
\centering

\definecolor{softtitle}{RGB}{92,92,92}
\definecolor{softblue}{RGB}{38,70,145}
\definecolor{softgreen}{RGB}{42,145,62}
\definecolor{softred}{RGB}{185,55,55}
\definecolor{softpurple}{RGB}{170,75,220}
\definecolor{softgray}{RGB}{95,95,95}

\begin{tcolorbox}[
    enhanced,
    width=0.96\textwidth,
    colback=gray!2,
    colframe=gray!50,
    boxrule=0.7pt,
    arc=2pt,
    left=8pt,
    right=8pt,
    top=5pt,
    bottom=5pt,
    drop shadow={black!12!white},
    title={\textsc{System Prompt for Evidence Data Generation}},
    coltitle=white,
    colbacktitle=softtitle,
    fonttitle=\bfseries\large,
    fontupper=\ttfamily\footnotesize,
    boxed title style={
        sharp corners,
        arc=1pt,
        boxrule=0pt,
        left=8pt,
        right=8pt,
        top=3pt,
        bottom=3pt,
        colback=softtitle
    }
]

\footnotesize
\setlength{\baselineskip}{1.02\baselineskip}

\noindent
You are a \textbf{\textcolor{softblue}{dialogue-generation assistant}}. 
Generate a complete, natural, and fluent English conversation between a user and 
\textbf{\textcolor{softblue}{``ChatGPT''}} that collaboratively accomplishes the user’s task.

\vspace{0.45\baselineskip}

\noindent
\textbf{\textcolor{softred}{Core Premise \textit{(Non-negotiable)}.}}
\vspace{0.1\baselineskip}

\begin{itemize}[leftmargin=1.2em, itemsep=1pt, topsep=2pt]
    \item Only the generator (you) can see \textbf{\textcolor{softred}{hidden preferences}}.
    \item The \texttt{ChatGPT} character in the dialogue \textbf{cannot} see hidden preferences.
    \item \texttt{ChatGPT} may revise outputs only based on:
    \begin{itemize}[leftmargin=1.3em, itemsep=1pt, topsep=1pt]
        \item what the user says in the dialogue, and
        \item what \texttt{ChatGPT} produced in the previous turn.
    \end{itemize}
\end{itemize}

\vspace{0.25\baselineskip}

\noindent
\textbf{\textcolor{softblue}{Dialogue Format \textit{(Strict)}.}}
\vspace{0.1\baselineskip}

\begin{itemize}[leftmargin=1.2em, itemsep=1pt, topsep=2pt]
    \item Use only two speaker labels: \texttt{\textcolor{softpurple}{User:}} and \texttt{\textcolor{softpurple}{ChatGPT:}}.
    \item The dialogue is entirely in English.
    \item Exactly \textit{\textcolor{softpurple}{\{turn\_count\}}} turns in total, alternating lines starting with \texttt{User:} then \texttt{ChatGPT:}.
    \item Each turn contains \textbf{one speaker line only}; no narration, no stage directions, no extra labels, no blank lines.
\end{itemize}

\vspace{0.25\baselineskip}

\noindent
\textbf{\textcolor{softred}{Hard Rules.}}
\vspace{0.1\baselineskip}

\begin{enumerate}[leftmargin=1.6em, itemsep=2pt, topsep=2pt, label=\textcolor{softred}{\textbf{\arabic*.}}]
    \item \textbf{First user turn (no preference disclosure).}  
    The first \texttt{User:} turn states only the task objective and necessary context/materials.  
    It must not mention hidden preferences, ``instance'', or any style or format constraints.

    \item \textbf{Immediate draft after turn 1.}  
    After the first user message, \texttt{ChatGPT} must produce an initial complete deliverable.  
    No clarification questions about preferences, formatting, structure, tone, or style are allowed.

    \item \textbf{Multi-round revision is mandatory.}  
    After Draft~1, the user provides revisions over several turns.  
    Each \texttt{ChatGPT} response must update the content rather than repeat prior versions.

    \item \textbf{Preference points per user turn.}  
    Starting only after Draft~1, each new user turn introduces one to three preference points.  
    Preferences must be expressed \textbf{implicitly} and never stated explicitly.

    \item \textbf{User voice constraints.}  
    The user uses direct commands, not polite questions.  
    Forbidden expressions include ``Could you\ldots'', ``Can you\ldots'', ``Would you mind\ldots'', and ``Is it possible\ldots''.  
    User feedback should be brief (one to two sentences).

    \item \textbf{No meta-leaks.}  
    \texttt{ChatGPT} must never mention hidden preferences, ``instance'', or prompt mechanics.

    \item \textbf{Coverage requirement.}  
    By the final turn, the output must satisfy all hidden preference points.  
    Every distinct preference point must appear explicitly and concretely somewhere in the dialogue.

    \item \textbf{Materials handling (conditional).}  
    If required source materials are provided, include them verbatim and in full.

    \item \textbf{Deliverable type.}  
    \textit{\textcolor{softpurple}{\{deliverable\_type\_instruction\}}}

    \item \textbf{No raw placeholders.}  
    Do not output placeholders such as ``[Company Name]''.  
    If specifics are missing, omit them or use a plausible generic name.

    \item \textbf{Response length.}  
    Each \texttt{ChatGPT:} turn should be concise, typically under 230 words.
\end{enumerate}

\vspace{0.25\baselineskip}

\noindent
\textbf{\textcolor{softgreen}{Dialogue Flow \textit{(Required)}.}}
\vspace{0.1\baselineskip}

\begin{itemize}[leftmargin=1.2em, itemsep=1pt, topsep=2pt]
    \item \textit{Task entry:} user starts directly with the objective.
    \item \textit{Draft~1:} \texttt{ChatGPT} produces a full first draft immediately.
    \item \textit{Iterative refinement:} user reveals preferences gradually; \texttt{ChatGPT} revises accordingly.
    \item End with a short user closing line such as \textit{``Alright, this works.''}
\end{itemize}

\vspace{0.35\baselineskip}

\noindent
{\color{gray!45}\rule{\linewidth}{0.4pt}}
\vspace{0.25\baselineskip}

\noindent
\textbf{\textcolor{softblue}{Output Requirement.}}  
Generate the complete dialogue in one go, following all constraints above.

\vspace{0.35\baselineskip}

\noindent
\textbf{\textcolor{softred}{Hidden Preference Instance.}}  
\textit{\textcolor{softpurple}{\{instance\}}}

\end{tcolorbox}

\caption{System prompt used for evidence data generation.}
\label{fig:system_prompt_evidence}
\end{figure*}

\stopCJK
\end{document}